\documentclass[lettersize,journal]{IEEEtran}
\usepackage{amsmath,amsfonts}
\usepackage{algorithmic}
\usepackage{algorithm}
\usepackage{array}
\usepackage[caption=false,font=normalsize,labelfont=sf,textfont=sf]{subfig}
\usepackage{textcomp}
\usepackage{stfloats}
\usepackage{url}
\usepackage{verbatim}
\usepackage{graphicx}
\usepackage{cite}
\usepackage{multirow}
\usepackage{threeparttable}
\usepackage{booktabs}
\begin{document}

\title{MSHyper: Multi-Scale Hypergraph Transformer for Long-Range Time Series Forecasting}

\author{Zongjiang Shang, Ling Chen, Binqing Wu, Dongliang Cui

\thanks{Corresponding author: Ling Chen.}
\thanks{This work was supported by the Science Foundation of Donghai Laboratory (Grant No. DH-2022ZY0013).}
\thanks{Zongjiang Shang, Ling Chen, Binqing Wu, and Dongliang Cui are with the State Key Laboratory of Blockchain and Data Security, College of Computer Science and Technology, Zhejiang University, Hangzhou 310027, China (e-mail: \{zongjiangshang, lingchen, binqingwu, runnercdl\}@cs.zju.edu.cn).}}


\markboth{Journal of \LaTeX\ Class Files,~Vol.~xx, No.~xx, xx~2024}%
{Shell \MakeLowercase{\textit{et al.}}: A Sample Article Using IEEEtran.cls for IEEE Journals}


\maketitle

\begin{abstract}
Demystifying the interactions between temporal patterns of different scales is fundamental to precise long-range time series forecasting. However, previous works lack the ability to model high-order interactions. To promote more comprehensive pattern interaction modeling for long-range time series forecasting, we propose a \textbf{\underline{M}}ulti-\textbf{\underline{S}}cale \textbf{\underline{Hyper}}graph Transformer (MSHyper) framework. Specifically, a multi-scale feature extraction (MFE) modlue is introduced to map the input sequence into multi-scale embeddings.  Then, a multi-scale hypergraph is designed to provide foundations for modeling high-order pattern interactions and a hyperedge graph is built to enhance hypergraph modeling. In addition, a tri-stage message passing (TMP) mechanism is introduced to aggregate pattern information and learn the interaction strength between temporal patterns of different scales. Extensive experimental results on eight real-world datasets demonstrate that MSHyper achieves the state-of-the-art (SOTA) performance across various settings.
\end{abstract}


\section{Introduction}
 \IEEEPARstart{T}{ime} series forecasting has demonstrated its wide applications across many fields (e.g., energy consumption planning \cite{TKDE2,TKDE3}, traffic and economics prediction \cite{TKDE1,ICDE1,TKDE5,WSDM2}, and disease propagation forecasting \cite{KDD2,KDD1,TOIS1,SIGMOD1}. In these real-world applications, how to use a substantial amount of previous time-series data and extend the forecasting horizon into the far future (i.e., long-range time series forecasting) is quite meaningful, as it can help decision makers to make schedule planning and optimize resource allocation.
 
Many time series demonstrate complex and diverse temporal patterns of different scales \cite{robustperiod,Quatformer,MAGNN,SIGMOD,TKDE4,WeatherGNN}. For example, due to periodic human activities, traffic occupation and electricity consumption show clear daily and weekly patterns. Considering the interactions between these temporal patterns often leads to more accurate forecasting results than analyzing each pattern separately. For example, the morning rush hour on the first workday after a holiday tends to be more congested (the interactions between daily and weekly patterns), while the evening rush hour on the last workday before a long holiday starts earlier. Therefore, how to model complex temporal patterns of different scales and their interactions is a fundamental problem in long-range time series forecasting.

To model temporal patterns of different scales and their interactions, traditional methods (e.g., seasonal ARIMA \cite{ARIMA} and Prophet \cite{Prophet}), use decomposition with heuristic priors to obtain temporal patterns of different scales, but cannot model complex non-linear dependencies of time series. Recently, deep neural networks have demonstrated superiority in capturing non-stationary and non-linear dependencies. Temporal convolutional networks (TCNs) \cite{TCNs,TimesNet}, recurrent neural networks (RNNs) \cite{DeepAR,TKDE6,TKDE7}, graph neural networks (GNNs) \cite{MTGNN,MSGNet,VLDB1,VLDB2}, and Transformers \cite{Informer,Autoformer,Quatformer,WSDM1} have been used for time series forecasting. To model temporal patterns of different scales, multi-scale Transformer-based methods \cite{LogTrans} attempt to build sub-sequences of different scales from the original input sequence but ignore the interactions between temporal patterns of different scales. To address this issue, recent multi-scale Transformer-based methods \cite{pyraformer,Triformer} introduce special structures (e.g., pyramidal structures) between sub-sequences of different scales. These structures model temporal dependencies within a sub-sequence through intra-scale edges, and model the interactions between temporal patterns of different scales through inter-scale edges.

However, these methods only use edges to model pairwise interactions and lack the ability to model high-order interactions (i.e., the simultaneous interactions between multiple temporal patterns). In reality, temporal patterns of different scales co-exist and exhibit high-order interactions, e.g., the peak household electricity consumption during summer weekend afternoons (high-order interactions between daily, weekly, and monthly patterns), as well as the high but relatively stable household electricity consumption during winter weekends.

To address the above issue, we propose MSHyper, a \textbf{\underline{M}}ulti-\textbf{\underline{S}}cale \textbf{\underline{Hyper}}graph Transformer framework for long-range time series forecasting. MSHyper aggregates the input sequence into sub-sequences of different scales, and models high-order interactions between temporal patterns of different scales by building multi-scale hypergraph structures. To the best of our knowledge, MSHyper is the first work that incorporates hypergraph modeling into long-range time series forecasting. The main contributions are as follows:

\begin{itemize}
\item{We propose a hypergraph and hyperedge graph construction (H-HGC) module that builds the hypergraph according to the temporal proximity rules, which can model intra-scale, inter-scale, and mixed-scale high-order interactions between temporal patterns. In addition, by treating hyperedges as nodes and building edges based on the sequential relationship and association relationship between nodes, H-HGC also builds the hyperedge graph to enhance hypergraph modeling.}

\item{We propose a tri-stage message passing (TMP) mechanism that has three message passing phases: Node-hyperedge, hyperedge-hyperedge, and hyperedge-node, which can aggregate pattern information and learn the interaction strength between temporal patterns of different scales.}

\item{We conduct extensive experiments on eight real-world time series datasets, and experimental results demonstrate that MSHyper achieves the state-of-the-art (SOTA) performance
across various settings.}
\end{itemize}

\section{Related Work}
In this section, we provide a brief review of the related work, including methods for time series forecasting, multi-scale Transformers, and hypergraph neural networks related studies.
\subsection{Methods for Time Series Forecasting}
Time series forecasting methods can be roughly divided into statistical methods and deep neural network-based methods. Statistical methods (e.g., ARIMA \cite{ARIMA} and Prophet \cite{Prophet}) follow arbitrary yet simple assumptions, and fail to model complicated temporal dependencies. Recently, deep neural networks show superiority in modeling complicated temporal dependencies \cite{LogTrans,MTGNN}. TAMS-RNNs \cite{TAMS-RNN} obtains the periodic temporal dependencies through multi-scale recurrent neural networks (RNNs) with different update frequencies. LSTNet \cite{LSTNet} utilizes recurrent-skip connections in combination with convolutional neural networks (CNNs) to capture long- and short-term temporal dependencies. Transformer-based methods \cite{survey1,survey2,Reformer,Informer,Autoformer} take advantage of the attention mechanism and achieve great access in modeling long-range temporal dependencies. Reformer \cite{Reformer} approximates the attention value through local-sensitive hashing (LSH), Informer \cite{Informer} obtains the dominant query by calculating the KL-divergence to realize the approximation of self-attention, and Autoformer \cite{Autoformer} introduces an auto-correlation mechanism to operate at the level of sub-sequences. However, the above methods struggle to obtain the complex temporal patterns of long-range time series.

\subsection{Multi-Scale Transformers} 
Multi-scale or hierarchical Transformers have been proposed in different fields (e.g.,  computer vision \cite{MVL,ViT,pyramidVT}, natural language processing \cite{ETC,Star-transformer,multi-scale}, and time series forecasting \cite{Quatformer,FEDformer,scaleformer}). Multi-scale ViT \cite{MVL} realizes image recognition by combining multi-scale image feature embeddings and Transformer. Star-Transformer \cite{Star-transformer} models intra-scale and inter-scale information interactions by introducing the global node embedding.
To address the limitations in the expressiveness of a single global node, ETCformer \cite{ETC} carries out information interactions between the global and local nodes by introducing a set of global node embeddings and setting fixed-length windows. To further extend the ability to model the interactions between temporal patterns of different scales, Pyraformer \cite{pyraformer} extends the two-layer structure into multi-scale embeddings, and models the interactions between nodes of different scales through a pyramid graph. Crossformer \cite{Crossformer} combines a two-stage attention with a hierarchical encoder-decoder architecture to capture cross-time and cross-dimension interactions. MSGNet \cite{MSGNet} combines an adaptive graph convolution and a temporal multi-head attention mechanism to capture multi-scale inter-series correlations. However, existing methods only model pairwise interactions between nodes, ignoring high-order interactions between temporal patterns of different scales.

\subsection{Hypergraph Neural Networks}
Hypergraph neural networks (HGNNs) have been proven to be capable of modeling high-order interactions, which have been applied to various fields (e.g., visual object recognition \cite{RC}, sequential recommendation \cite{MBHT,SIGIR}, trajectory prediction \cite{groupnet}, stock selection \cite{stockselection,ICDE2}, and citation network classification \cite{HAHC, TKDE8}). HGNN \cite{HGNN} and HyperGCN \cite{hypergcn} are the first works to apply graph convolution to hypergraphs, which demonstrate the superiority of hypergraphs over ordinary graph neural networks (GNNs) in modeling high-order interactions. MBHT \cite{MBHT} combines hypergraphs with a Transformer framework for the sequential recommendation, which leverages hypergraphs to capture high-order user-item interaction patterns. GroupNet \cite{groupnet} uses multi-scale hypergraphs for trajectory prediction, which leverages topology inference and representation learning to capture the agent patterns and their high-order interactions. STHAN-SR \cite{stockselection} leverages hypergraph and temporal Hawkes attention mechanism for stock selection, which leverages hypergraph network architecture to model inter stock relations of varing types and degrees.

Considering the ability of HGNNs in high-order interaction modeling, we propose a multi-scale hypergraph Transformer framework to model high-order interactions between temporal patterns of different scales. Specifically, a H-HGC module is introduced to build the hypergraph and hyperedge graph. In addition, a TMP mechanism is proposed to aggregate pattern information and learn the interaction strength between temporal patterns of different scales by three message passing phases.
\section{Preliminaries}
In this section, we first provide the definition of hypergraph and then formulate the problem.
\subsection{Hypergraph}
A hypergraph is defined as $\mathcal{G}=\{\mathcal{V},\mathcal{E}\}$, where $\mathcal{E}=\{e_1,e_2,...,e_M\}$ is the hyperedge set and $\mathcal{V}=\{v_1,v_2,...,v_N\}$ is the node set. Each hyperedge $e_m\in\mathcal{E}$ connects a set of nodes in $\mathcal{V}$.
The hypergraph $\mathcal{G}$ can be denoted as an incidence matrix $\mathbf{H}\in\mathbb{R}^{N\times M}$, where $\mathbf{H}_{nm}=1$ if the $n\text{th}$ node belongs to the $m\text{th}$ hyperedge, else $\mathbf{H}_{nm}=0$. The degree of the $n\text{th}$ node is defined as follows:
\begin{equation}
\label{1}
\mathbf{D}_{v_n}=\sum_{m=1}^M\mathbf{H}_{nm}.
\end{equation}
The degree of the $m\text{th}$ hyperedge is defined as follows:
\begin{equation}
\label{2}
\mathbf{D}_{e_m}=\sum_{n=1}^N\mathbf{H}_{nm}.
\end{equation}
The results of node degrees and hyperedge degrees are stored in diagonal matrices $\textbf{D}_\text{v}\in\mathbb{R}^{N\times N}$ and $\mathbf{D}_\mathrm{e}\in\mathbb{R}^{M\times M}$, respectively.
\subsection{Problem Statement}
The task of long-range time series forecasting is to predict the future $H$ steps given the previous $T$ steps of observed values. Specifically, given the input sequence $\bold{X}_{1: T}^{\text{I}}=\left\{\boldsymbol{x}_t \mid \boldsymbol{x}_t \in \mathbb{R}^D, t \in[1, T]\right\}$, where $\boldsymbol{x}_t $ represents the values at time step $t$, and $D$ is the feature dimension, the prediction task is defined as follows:
\begin{equation}
\label{3}
\widehat{\bold{X}}_{T+1: T+H}^{\text{O}}=\mathcal{F}\left(\bold{X}_{1: T}^{\text{I}} ; \theta\right) \in \mathbb{R}^{H \times D},
\end{equation}
where $\widehat{\mathbf X}_{T+1:T+H}^\text{O}$ denotes the forecasting results, $\mathcal{F}$ denotes the mapping function, and $\theta$ denotes the learnable parameters of $\mathcal{F}$.
\section{Methodology}
In this section, we give the description of the proposed MSHyper. We first give the framework of MSHyper, and then present the detailed descriptions of each module. In addition, we introduce the detailed process of the tri-stage message passing (TMP) mechanism based on the framework of MSHyper.
\begin{figure*}[]
\includegraphics[width=0.55\textwidth]{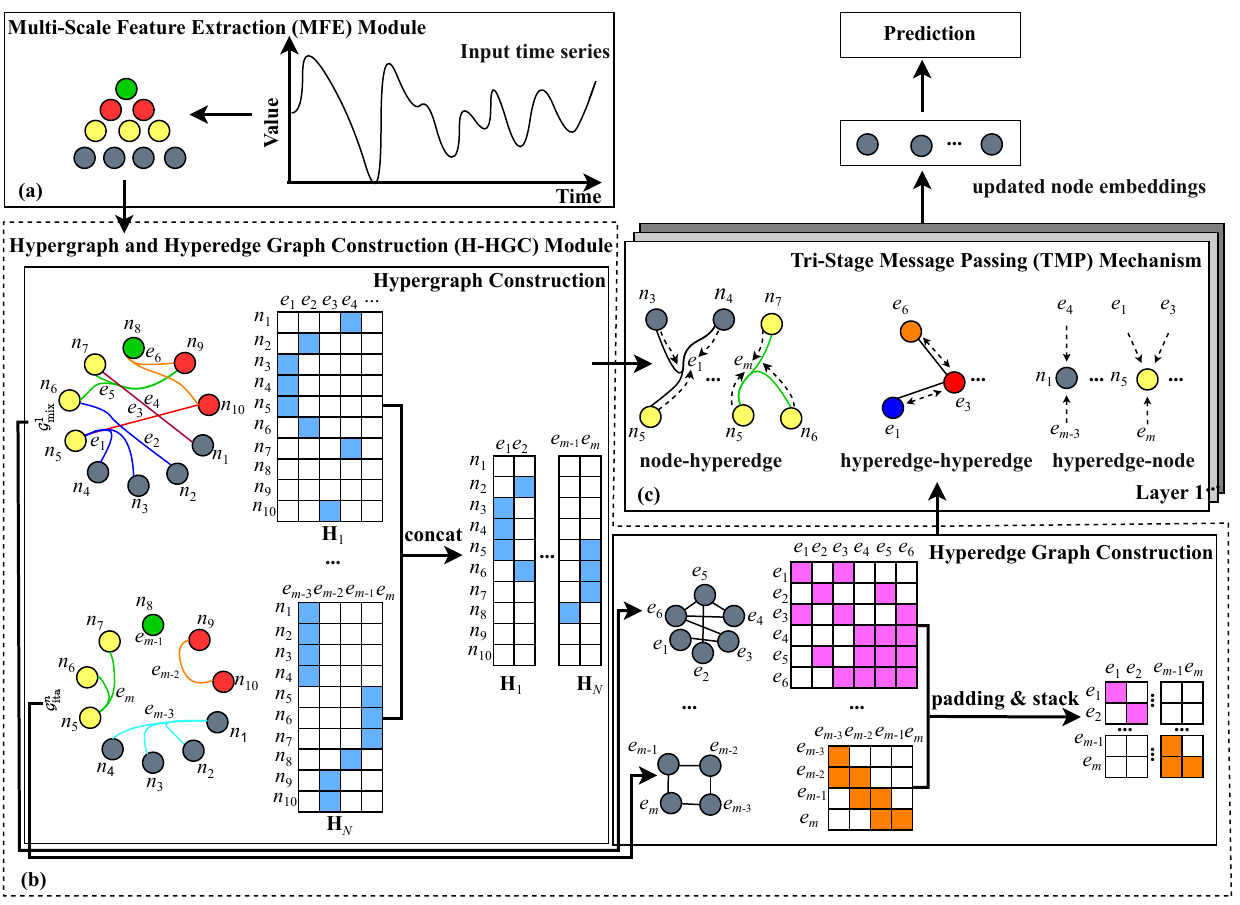}
\centering
\caption{The framework of MSHyper, which consists of three parts: (a) The HFE module, which maps the input sequence into hierarchical embeddings. (b) The H-HGC module, which provides foundations for modeling high-order interactions between temporal patterns by building the hypergraph and the hyperedge graph. (c) The TMP mechanism, which aggregates pattern information and learns the interaction strength between temporal patterns of different scales.}
\label{Figure_1}
\end{figure*}
\subsection{Framework}
As mentioned above, the core of MSHyper is to build multi-scale hypergraph structures, which can explicitly model high-order interactions between temporal patterns of different scales. To accomplish this goal, we first map the input sequence into multi-scale embeddings through the multi-scale feature extraction (MFE) module and then leverage the H-HGC module to build the hypergraph and hyperedge graph. Finally, we employ the TMP mechanism to aggregate pattern information and learn the interaction strength between temporal patterns of different scales. Figure \ref{Figure_1} illustrates the framework of MSHyper.

\subsection{Multi-Scale Feature Extraction Module}
To get the feature embeddings of different scales, we first map the input sequence into multi-scale embeddings. We use $\mathbf{X}^s=\{\boldsymbol{x}_t^s|\boldsymbol{x}_t^s\in\mathbb{R}^D,t\in[1,h^s]\}$ to represent the sub-sequence at scale $s$, where $s=1,...,S$ denotes the scale index, and $S$ is the total number of scales. $h^{s}=\left\lfloor\frac{h^{s-1}}{l^{s-1}}\right\rfloor\text{s.t.}~{} s\geq2$ is the horizon at scale $s$ and $l^{s-1}$ denotes the size of the aggregation window at scale $s-1$. $\mathbf{X}^1=\mathbf{X}^\text{I}_{1:T}$ is the raw input sequence and the aggregation process is defined as follows:
\begin{equation}
\label{4}
\mathbf{X}^{s+1}=Aggregation(\mathbf{X}^s;\theta^s)\in\mathbb{R}^{h^{s+1}\times D},
\end{equation}
where $Aggregation$ is the aggregation function (e.g., 1D convolution or average pooling), and $\theta^s$ denotes the learnable parameters of the aggregation function at scale $s$.
\subsection{Hypergraph and Hyperedge Graph Construction  Module}
Long-range time series data contain a lot of noise and are non-stationary \cite{Autoformer}. When modeling high-order interactions between temporal patterns without prior knowledge constraints, the model tends to learn spurious interactions and lacks interpretability. To comprehensively model high-order interactions between temporal patterns of different scales, we build the hypergraph and the hyperedge graph separately by the H-HGC module to model intra-scale, inter-scale, and mixed-scale high-order pattern interactions.
\subsubsection{Hypergraph Construction}

\begin{figure}[!t]
\centering
\includegraphics[width=0.5\textwidth]{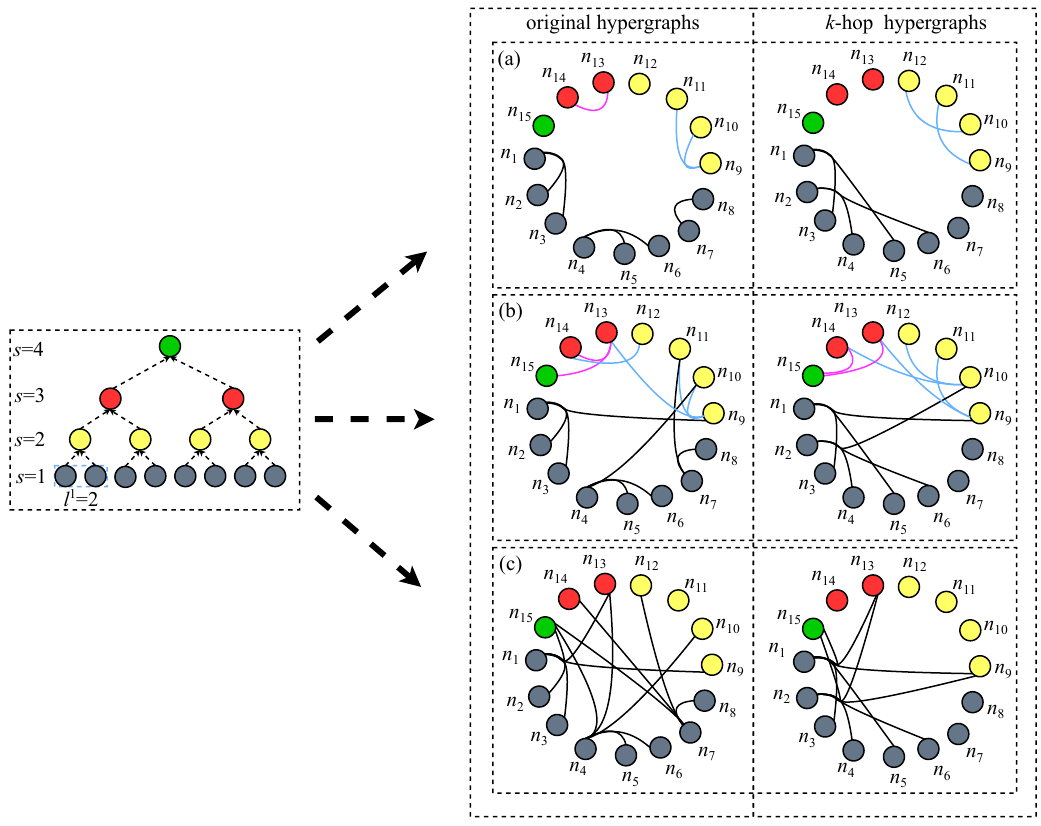}
\caption{Hypergraph construction. (a), (b), and (c) represent the intra-scale hypergraph, inter-scale hypergraph, and mixed-scale hypergraph, respectively. In addition to the original connections, we also aggregate information from different ranges of neighbors using $k$-hop connections.}
%
\label{Figure_2}
\end{figure}

Despite existing methods \cite{pyraformer,graphwave} being capable of modeling the interactions between temporal patterns of different scales through graph structures, we argue that these methods still face two limitations: (1) These methods fail to model high-order interactions between temporal patterns. (2) These methods limit interactions to the neighboring nodes and thus are incapable of capturing the interactions between nodes that are far away but still show correlations. To address these problems, we build the hypergraph structures according to the temporal proximity rules. As shown in Figure \ref{Figure_2}, on the one hand, we build different types of hypergraphs to model intra-scale, inter-scale, and mixed-scale high-order interactions. On the other hand, we build the $k$-hop hypergraph under each type of hypergraph to aggregate information from different ranges of neighbors.

\textbf{Intra-Scale Hypergraph.} As shown in Figure \ref{Figure_2}(a), in order to capture high-order interactions between intra-scale temporal patterns, we construct the intra-scale hypergraph $\mathcal{G}_{\text{ita}}$, which contains two hypergraphs (i.e., the original intra-scale hypergraph $\mathcal{G}_{\text{o, ita}}$ and $k$-hop intra-scale hypergraph $\mathcal{G}_{\text{k, ita}}$). We use $\mathcal{G}_{\text{o, ita}}=\{\mathcal{V},\mathcal{E}_{\text{o, ita}}\}$ to represent the original intra-scale hypergraph, where $\mathcal{E}_{\text{o, ita}}=\{\mathcal{E}_{\text{o, ita}}^s\}_{s\in\{1,...,S\}}$ contains original intra-scale hyperedge sets of different scales. The $i\text{th}$ hyperedge of scale $s$ $e_i^s\in\mathcal{E}_{\text{o, ita}}^s$ is defined as follows:
\begin{equation}
\label{5}
e_i^s=\left\{v_{\epsilon}^s,\forall v_j^s\in\mathcal{N}(v_{\epsilon}^s)\right\}\text{s.t.}~{}0<j-\epsilon\leq H_s,
\end{equation}
where $\epsilon=(i-1)H_s+1$ is the starting node index under the $i\text{th}$ hyperedge of scale $s$ based on original connections. $H_s$ is the number of nodes connected by each hyperedge of scale $s$ and $\mathcal{N}(v_{\epsilon}^s)$ is the neighboring nodes connected to node $v_{\epsilon}^s$. Meanwhile, we use $\mathcal{G}_{\text{k, ita}}=\{\mathcal{V},\mathcal{E}_{\text{k, ita}}\}$ to represent the $k$-hop intra-scale hypergraph, where $\mathcal{E}_{\text{k, ita}}=\{\mathcal{E}_{\text{k, ita}}^s\}_{s\in\{1,...,S\}}$ contains $k$-hop intra-scale hyperedge sets of different scales. The $i\text{th}$ hyperedge of scale $s$ based on $k$-hop connections $e^s_{i,k}\in\mathcal{E}^s_{\text{k, ita}}$ is defined as follows:
\begin{equation}
\label{6}
e_{i,k}^s=\{v_d^s,v_{d+k}^s,\ldots,v_{d+(H_s-1)k}^s\},
\end{equation}
where $d$ is the starting node index under the $i\text{th}$ hyperedge of scale $s$ based on $k$-hop connections, which is defined as follows:
\begin{equation}
\label{7}
d=\left\lfloor\frac{i-1}{k}\right\rfloor\times H_sk+(i-1)\%k+1,
\end{equation}
where $k$ is the temporal distance between two neighboring nodes. 
The intra-scale hypergraph $\mathcal{G}_{\text{ita}}=\{\mathcal{V},\mathcal{E}_{\text{ita}}\}$ based on the original intra-scale hypergraph and the $k$-hop intra-scale hypergraph is defined as follows:
\begin{equation}
\label{8}
\mathcal{G}_{\text{ita}}=concat(\mathcal{G}_{\text{o, ita}},\mathcal{G}_{\text{k, ita}}),
\end{equation}
where $concat$ denotes the concatenation operation. High-order interactions between temporal patterns not only exist between intra-scale temporal patterns, but also between inter-scale temporal patterns (e.g., the interactions between hourly and daily patterns, and the interactions between daily and weekly patterns). In addition, there are temporal pattern interactions across all scales (e.g., the interactions between hourly, daily, and weekly patterns). Therefore, we design the inter-scale hypergraph and mixed-scale hypergraph.

\textbf{Inter-Scale Hypergraph.} As shown in Figure \ref{Figure_2}(b), the inter-scale hypergraph
$\mathcal{G}_{\text{ite}}=\{\mathcal{V},\mathcal{E}_{\text{ite}}\}$ is obtained  by concatenating the original inter-scale hypergraph and $k$-hop inter-scale hypergraph. We use $\mathcal{G}_{\text{o, ite}}=\{\mathcal{V},\mathcal{E}_{\text{o, ite}}\}$ to represent the original inter-scale hypergraph, where $\mathcal{E}_{\text{o, ite}}=\{\mathcal{E}_{\text{o, ite}}^s\}_{s\in\{1,...,S\}}$ contains original inter-scale hyperedge sets of different scales. The $i\text{th}$ hyperedge of scale $s$ $e^s_i\in\mathcal{E}^s_{\text{o, ite}}$ is defined as follows:
\begin{equation}
\label{9}
e_i^s=\left\{v_{\lceil\epsilon/{l^s}\rceil}^{s+1},v_{\epsilon}^s,v_{\epsilon+1}^s,\ldots,v_{\epsilon+H_s}^s\right\},
\end{equation}
where $l^s$ denotes the aggregation window size at scale $s$. Then, we use $\mathcal{G}_{\text{k, ite}}=\{\mathcal{V},\mathcal{E}_{\text{k, ite}}\}$ to represent the $k$-hop inter-scale hypergraph, where $\mathcal{E}_{\text{k, ite}}=\{\mathcal{E}_{\text{k, ite}}^s\}_{s\in\{1,...,S\}}$ contains $k$-hop inter-scale hyperedge sets of different scales. The $i\text{th}$ hyperedge of scale $s$ based on $k$-hop connections $e^s_{i,k}\in\mathcal{E}^s_{\text{k,ite}}$ is defined as follows:
\begin{equation}
\label{10}
e^s_{i,k}=\{v_{\lceil d/l^s\rceil}^{s+1},v_d^s,v_{d+k}^s,\dots, v_{d+(H_s-1)k}^s\}.
\end{equation}
The inter-scale hypergraph $\mathcal{G}_{\text{ite}}=\{\mathcal{V},\mathcal{E}_{\text{ite}}\}$ based on the original inter-scale hypergraph and the $k$-hop inter-scale hypergraph is defined as follows:
\begin{equation}
\label{11}
\mathcal{G}_{\text{ite}}=concat(\mathcal{G}_{\text{o, ite}},\mathcal{G}_{\text{k, ite}}).
\end{equation}

\textbf{Mixed-Scale Hypergraph.} As shown in Figure \ref{Figure_2}(c), the mixed-scale hypergraph 
$\mathcal{G}_{\text{mix}}=\{\mathcal{V},\mathcal{E}_{\text{mix}}\}$ is obtained  by 
 concatenating the original mixed-scale hypergraph and  $k$-hop mixed-scale hypergraph. We use $\mathcal{G}_{\text{o, mix}}=\{\mathcal{V},\mathcal{E}_{\text{o, mix}}\}$ to represent the original mixed-scale hypergraph, where $\mathcal{E}_{\text{o, mix}}$ denotes the original mixed-scale hyperedge set. The $i\text{th}$ hyperedge $e_i\in\mathcal{E}_{\text{o, mix}}$ is defined as follows:
\begin{equation}
\label{12}
e_i=\left\{v_{\delta_S}^S,v_{\delta_{S-1}}^{S-1},\dots,v_{\delta_{1}}^1,v_{\delta_1+1}^1,\dots,v_{\delta_1+H-1}^1\right\},
\end{equation}
where $\delta_s=\lceil\epsilon/(1\times\prod_{\alpha=2}^s l^\alpha)\rceil$ is the starting node index of scale $s$ under the $i$th hyperedge based on original connections.
Then, we use $\mathcal{G}_{\text{k, mix}}=\{\mathcal{V},\mathcal{E}_{\text{k, mix}}\}$ to represent the $k$-hop mixed-scale hypergraph, where $\mathcal{E}_{\text{k, mix}}$ denotes the $k$-hop mixed-scale hyperedge set. 
The $i\text{th}$ hyperedge based on $k$-hop connections $e_{i,k}\in\mathcal{E}_{\text{k, mix}}$ is defined as follows:
\begin{equation}
\label{13}
\resizebox{.87\linewidth}{!}{$
e_{i,k}=\left\{v_{\delta_{S,k}}^S,v_{\delta_{S-1,k}}^{S-1},\ldots,v_{\delta_1}^1,v_{\delta_1+k}^1,\ldots,v_{\delta_1+(H-1)k}^1\right\}$},
\end{equation}
where $\delta_{s,k}=\lceil{d/(1\times\prod_{\alpha=2}^s l^\alpha)}\rceil$ is the starting node index of scale $s$ under the $i$th hyperedge based on $k$-hop connections. 

The mixed-scale hypergraph $\mathcal{G}_{\text{mix}}=\{\mathcal{V},\mathcal{E}_{\text{mix}}\}$ based on the original mixed-scale hypergraph and the $k$-hop mixed-scale hypergraph is defined as follows:
\begin{equation}
\label{14}
\mathcal{G}_{\text{mix}}=concat(\mathcal{G}_{\text{o,mix}},\mathcal{G}_{\text{k,mix}}).
\end{equation}

The hypergraph construction module generates the hypergraph $\mathcal{G}=\{\mathcal{V},\mathcal{E}\}$ by considering the intra-scale hypergraph, the inter-scale hypergraph, and the mixed-scale hypergraph, which is defined as follows:
\begin{equation}
\label{15}
\mathcal{G}=concat(\mathcal{G}_{\text{ita}},\mathcal{G}_{\text{ite}},\mathcal{G}_{\text{mix}}).
\end{equation}
\subsubsection{Hyperedge Graph Construction}
After building the hypergraph, to enhance hypergraph modeling, we build the hyperedge graph to model the interactions between hyperedges. In GNNs, Chen et al.\cite{linegraph} proposes the concept of the line graph to enhance graph modeling through edge interactions. However, since the line graph is unweighted and considers two edges to be correlated only if the target node of one edge is the source node of the other edge, it lacks in characterizing various edge interaction patterns and is not suitable for modeling hyperedge interactions that connect multiple nodes. Thus, we build the hyperedge graph by representing hyperedges in $\mathcal{G}$ as nodes and considering the sequential relationship and association relationship. We define the hyperedge graph as $G=(V,E,\mathbf{A})$, where $V=\{v_{e_i}|v_{e_i}\in\mathcal{E}\}$ is the node set of $G$ and $E=\{(v_{e_i},v_{e_j})|v_{e_i},v_{e_j}\in\mathcal{E}\}$ is the edge set of $G$.  $\mathbf{A}$ is the weighted adjacency matrix defined based on the sequential relationship and association relationship between hyperedges.

\begin{figure}[]
\centering
\includegraphics[width=0.5\textwidth]{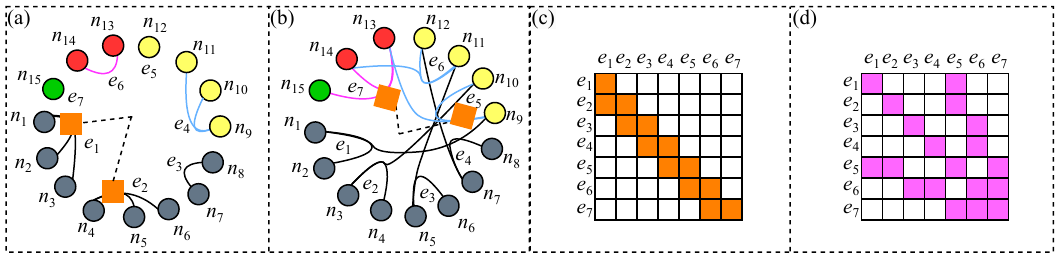}
\caption{Hyperedge graph construction. (a) and (b) are the sequential relationship and association relationship, respectively. (c) and (d) are the constructed adjacency matrix based on the sequential relationship and association relationship, respectively. }
\label{Figure_3}
\end{figure}

\textbf{Sequential Relationship.} Within the intra-scale hypergraph, two hyperedges that connect nodes with temporal order exhibit the sequential relationship. For example, as shown in Figure \ref{Figure_3}(a), hyperedge $e_1$ connects the previous three nodes, and hyperedge $e_2$ connects the next three nodes. Since long-range time series is a collection of data points arranged in chronological order, intuitively, changes (e.g., increase, decrease, or fluctuation) in the values of nodes connected by $e_1$ may influence the values of nodes connected by $e_2$. We use $G_{\text{sr}}=(V_{\text{sr}}, E_{\text{sr}},\mathbf{A}_{\text{sr}})$ to represent the hyperedge graph constructed based on the sequential relationship, where $V_{\text{sr}}=\{v_{{e}_{\text{sr}}}|v_{{e}_{\text{sr}}}\in\mathcal{E}_{\text{ita}}\}\in\mathbb{R}^{{D}_{\text{sr}}\times{D}}$ is the node set and $D_{\text{sr}}$ is the number of intra-scale hyperedges. $E_{\text{sr}}=\{(v_{{e}_{i,\text{sr}}},v_{{e}_{j,\text{sr}}})|v_{{e}_{i,\text{sr}}},v_{{e}_{j,\text{sr}}}\in\mathcal{E}_{\text{ita}},0\leq i-j\leq1\}$ denotes the edge set constructed based on the sequential relationship. As shown in Figure \ref{Figure_3}(c), the adjacency matrix constructed based on the sequential relationship $\mathbf{A}_{\text{sr}}\in\mathbb{R}^{D_{\text{sr}}\times D_{\text{sr}}}$ is defined as follows:
\begin{equation}
\label{16}
\resizebox{.87\linewidth}{!}{$
\mathbf{A}_{\text{sr}}=\{\mathbf{A}_{ij, \text{sr}}|\mathbf{A}_{ij, \text{sr}}=1\ \text{if}\ (v_{e_{i,\text{sr}}},v_{e_{j,\text{sr}}})\in E_{\text{sr}},\ \text{else}\ 0\}$}.
\end{equation}
\textbf{Association Relationship.} Within the inter-scale hypergraph and mixed-scale hypergraph, hyperedges with common nodes have an association relationship. For example, as shown in Figure \ref{Figure_3}(b), $e_5$ and $e_7$ share a common node $n_{13}$. The value changes of nodes connected by $e_5$ may influence the values of nodes connected by $e_7$ through the common node $n_{13}$. We use $G_{\text{ar}}=(V_{\text{ar}}, E_{\text{ar}},\mathbf{A}_{\text{ar}})$ to represent the hyperedge graph constructed based on the association relationship, where $V_{\text{ar}}=\{v_{e_{\text{ar}}}|v_{e_{\text{ar}}}\in\{\mathcal{E}_{\text{ite}},\mathcal{E}_{\text{mix}}\}\}\in\mathbb{R}^{D_{\text{ar}}\times D}$ is the node set of $G_{\text{ar}}$ and $D_{\text{ar}}$ is the number sum of inter-scale hyperedges and mixed-scale hyperedges. $E_{\text{ar}}=\{(v_{{e}_{i,\text{ar}}},v_{{e}_{j,\text{ar}}})|v_{{e}_{i,\text{ar}}},v_{{e}_{j,\text{ar}}}\in\{\mathcal{E}_{\text{ite}},\mathcal{E}_{\text{mix}}\},|v_{{e}_{i,\text{ar}}}\cap v_{{e}_{j,\text{ar}}}|\geq1\}$ denotes the edge set of $G_{\text{ar}}$. As shown in Figure \ref{Figure_3}(d), the adjacency matrix based on the association relationship $\mathbf{A}_{\text{ar}}\in\mathbb{R}^{D_{\text{ar}}\times D_{\text{ar}}}$ is defined as follows:
\begin{equation}
\label{17}
\resizebox{.87\linewidth}{!}{$
\mathbf{A}_{\text{ar}}=\{\mathbf{A}_{ij,\text{ar}}|\mathbf{A}_{ij,\text{ar}}=1\ \text{if}\ (v_{e_{i,\text{ar}}},v_{e_{j,\text{ar}}})\in E_{\text{ar}},\ \text{else}\ 0\}$}.
\end{equation}

The adjacency matrix $\mathbf{A}$ based on the above two relationships is defined as follows:
\begin{equation}
\label{18}
\mathbf{A}=\begin{bmatrix}\mathbf{A}_{\text{sr}}&\mathbf{\Gamma}_1\\ \mathbf{\Gamma}_2&\mathbf{A}_{\text{ar}}\end{bmatrix}\in\mathbb{R}^{D_\varphi\times D_\varphi},
\end{equation}
where $\boldsymbol{\Gamma}_1\in\mathbb{R}^{D_{\text{sr}}\times D_{\text{ar}}}$ and $\boldsymbol{\Gamma}_2\in\mathbb{R}^{D_{\text{ar}}\times D_{\text{sr}}}$ are matrices consisting entirely of zeros, and $D_\varphi$ is the sum of ${D}_\mathrm{\text{ar}}$ and ${D}_\mathrm{\text{sr}}$.

\subsection{Tri-Stage Message Passing Mechanism}
After building the hypergraph and hyperedge graph, to aggregate pattern information and learn the interaction strength between temporal patterns of different scales, we propose a TMP mechanism, which contains the node-hyperedge, hyperedge-hyperedge, and hyperedge-node phases.

\textbf{Node-Hyperedge Phase.}
Given the sequences based on the MFE module $\mathbf{X}=\{\mathbf{X}^{1},\mathbf{X}^{2},\ldots,\mathbf{X}^{S}\}\in\mathbb{R}^{N\times D}$, where $N$ denotes the number sum of input time steps and aggregated feature values of different scales. We first get the initialized node embeddings $\boldsymbol{\mathcal{V}}=f(\mathbf{X})\in\mathbb{R}^{N\times D}$,
where $f$ can be implemented by the multi-layer perceptron (MLP). As shown in Figure \ref{Figure_1}(c), we get the initialized hyperedge embeddings by the aggregation operation based on the hypergraph $\mathcal{G}$. Specifically, for the $i\text{th}$ hyperedge $e_i \in \mathcal{E}$, its initialized embedding is defined as follows:
\begin{equation}
\label{20}
\boldsymbol{v}_{e_i}=\sum\limits_{v_j\in \mathcal{N}(e_i)}\boldsymbol{v}_j\in\mathbb{R}^D,
\end{equation}
where $\mathcal{N}(e_i)$ denotes neighboring nodes connected by $e_i$.

\textbf{Hyperedge-Hyperedge Phase.}
After getting initialized hyperedge embeddings, we proceed to update their embeddings through the constructed hyperedge graph. Specifically, for the given initialized hyperedge embeddings $\mathbf{V}=\{\boldsymbol{v}_{e_i}|\boldsymbol{v}_{e_i}\in\mathcal{E}\}\in\mathbb{R}^{M\times D}$, we transform it into query $\widetilde{\mathbf{Q}}=\mathbf{VW}^{\text{q}}$, key $\widetilde{\mathbf K}=\mathbf{V}\mathbf{W}^{\text{k}}$, and value $\widetilde{\mathbf{V}}=\mathbf{V}\mathbf{W}^{\text{v}}$, where $\mathbf{W}^{\text{q}}$, $\mathbf{W}^{\text{k}}$, and $\mathbf{W}^{\text{v}}$ are learnable weight matrices. For the $i\text{th}$ row $\boldsymbol{\widetilde{q}}_i \in \widetilde{\mathbf{Q}}$, the updated hyperedge embedding $\widetilde{\boldsymbol{v}}_{e_i}\in\widetilde{\mathbf{V}}_\text{e}$ is defined as follows:
\begin{equation}
\label{21}
\begin{aligned}
&\widetilde{\boldsymbol{v}}_{e_i}=\sum_{j=1}^M\frac{exp(e_{ij})}{\Sigma_{\ell=1}^Mexp(e_{i\ell})}\widetilde{\boldsymbol{v}}_j  \\
&e_{ij} =\frac{\widetilde{\boldsymbol{q}}_i\widetilde{\boldsymbol{k}}_j^T}{\sqrt{D}}-(1-\mathbf{A}_{ij})\text{C}, 
\end{aligned}
\end{equation}
where $\boldsymbol{\widetilde{k}}_j^T$ denotes the transpose of the $j\text{th}$  row in $\widetilde{\mathbf{K}}$ and $\boldsymbol{\widetilde{v}}_j$ denotes the $j\text{th}$ row in $\widetilde{\mathbf{V}}$. $\mathbf{A}_{ij}\in\mathbf{A}$ is a binary value and $\text{C}$ is a large constant.

\textbf{Hyperedge-Node Phase.}
After obtaining the updated hyperedge embeddings, we update the node embeddings by considering all the related hyperedges. Considering the constructed hypergraph $\mathcal{G}$, we use the hypergraph convolution to update the node embeddings. Specifically, the symmetric normalized hypergraph Laplacian convolution is defined as follows:
\begin{equation}
\label{22}
\boldsymbol{\widetilde{\mathcal{V}}}=\sigma(\mathbf{D}_v^{-1/2}\mathbf{HD}_e^{-1}\mathbf{H}^T\mathbf{D}_v^{-1/2}\boldsymbol{{\mathcal{V}}}\mathbf{{P}})\in\mathbb{R}^{{N}\times{D}_c},
\end{equation}
where $\boldsymbol{\widetilde{\mathcal{V}}}$ is the output of the hypergraph convolution and ${D}_c$ is the output dimension. $\mathbf{P}\in\mathbb{R}^{D\times D_c}$ denotes the learnable parameters and $\sigma$ is the activation function (e.g., LeakyReLU and ELU). To capture the interaction strength of each node $v_i\in\boldsymbol{\mathcal{V}}$ and its related hyperedges, we dynamically update $\mathbf{H}$ using the node embedding and updated hyperedge embeddings, which is defined as follows: 
\begin{equation}
\label{23}
\mathbf{H}_{ij}^{\text{att}}=\frac{exp(\sigma(f_t[\boldsymbol{v}_i,\widetilde{\boldsymbol{v}}_{e_j}]))}{\sum_{k\in\mathcal{N}_i}exp(\sigma(f_t[\boldsymbol{v}_i,\widetilde{\boldsymbol{v}}_{e_k}])))},
\end{equation}
where $[.,.]$ denotes the concatenation operation of the node and its related hyperedges. $f_t$ is a trainable MLP, and $\mathcal{N}_i$ is the neighboring hyperedges connected to $v_i$, which can be accessed using the constructed hypergraph $\mathcal{G}$. Then, we use Equation \ref{22} to aggregate pattern information of different scales by replacing $\mathbf{H}$ with $\mathbf{H}^{\text{att}}$. The multi-head attention mechanism is also used to stabilize the training process, which is defined as follows:
\begin{equation}
\label{24}
\boldsymbol{\widetilde{\mathcal{V}}}=\bigoplus_{\jmath=1}^{\mathcal{J}}(\sigma(\mathbf{D}_v^{-1/2}\mathbf{H}_{\jmath}^{\mathrm{att}}\mathbf{D}_e^{-1}\mathbf{H}_\jmath^{\mathrm{att}^T}\mathbf{D}_v^{-1/2}\boldsymbol{\mathcal{V}}\mathbf{P}_\jmath)),
\end{equation}
where $\bigoplus$ is the aggregation function used for combing the outputs of multi-head (e.g., concatenation or average pooling). $\mathbf{H}^{\text{att}}_{\jmath}$ and $\mathbf{P}_{\jmath}$ are the enriched incidence matrix and the learnable weight matrix of the $j\text{th}$ head, respectively. $\mathcal{J}$ is the number of heads.

\begin{table}[]
\centering
  \caption{Dataset statistics.}
  \label{tab:addtable11}
  \resizebox{0.5\textwidth}{!}{
\begin{tabular}{lllll}
\hline
Datasets      & \#Variates & ACF values & Frequency  & Information    \\ \hline
Flight        & 7          & 0.73       & Hourly     & Flight         \\ 
Weather       & 21         & 0.35       & 10 minutes & Meteorological \\ 
ETTh1, ETTh2  & 7          & 0.46, 0.41 & Hourly     & Temperature    \\ 
ETTm1, ETTm2  & 7          & 0.39, 0.09  & 15 minutes & Temperature    \\ 
Electricity   & 321        & 0.84       & Hourly     & Electricity    \\ 
Exchange-Rate & 8          & 0.02       & Daily      & Economy        \\ \hline
\end{tabular}
}
\end{table}
\subsection{Objective Function}
After obtaining the updated node embeddings encoded by the multi-scale hypergraph, we concatenate the last node embeddings of each sub-sequence from different scales and then put them into a linear layer for prediction. We choose MSE as our objective function, which is defined as follows:
\begin{equation}
\label{25}
\mathcal L=\frac{1}{H}\left\|\widehat{\mathbf X}_{T+1:T+H}^\text{O}-\mathbf X_{T+1:T+H}^\text{O}\right\|_2^2,
\end{equation}
where $\widehat{\mathbf X}_{T+1:T+H}^\text{O}$ and $\mathbf X_{T+1:T+H}^\text{O}$ are forecasting results and ground truth, respectively.

\subsection{Complexity Analysis}
The time complexity of MSHyper consists of three main parts. For the node-hyperedge phase, the time complexity is $\mathcal{O}(M)$, where $M$ is the number of hyperedges. For the hyperedge-hyperedge phase, the time complexity is $\mathcal{O}(M^2)$. For the hyperedge-node phase, since $\mathbf{D}_v$ and $\mathbf{D}_e$ are diagonal matrices, and the sparse operation in \textit{torch\_geometric} of PyTorch, the time complexity is $\mathcal{O}(MN)$, where $N$ is the number of nodes. In practical operation, since the large aggregation window $l^{s}$ and a hyperedge can connect multiple nodes (large $H_{s}$), $M$ is smaller than $N$. As a result, the total complexity of MSHyper is bounded by $\mathcal{O}(MN)$.
\section{Experiments}
In this section, we first present experiments to verify the performance of MSHyper on eight public time series forecasting datasets. Then, we conduct ablation studies, case studies, and parameter studies to verify the effect of different module designs and parameter choices. In addition, we provide computation cost to verify the effectiveness of MSHyper.
\subsection{Datasets and Settings}

\textbf{Datasets.} We conduct experiments on eight commonly used long-range time series forecasting datasets, including \textit{ETT} (\textit{ETTh} and \textit{ETTm}), \textit{Electricity}, \textit{Weather}, \textit{Flight}, and \textit{Exchange-Rate} datasets. Table \ref{tab:addtable11} shows the summarized dataset statistics. The auto-correlation function (ACF) values is used to evaluate the correlation between a time series and its lagged values \cite{ACF}. Higher ACF values typically indicates greater predictability. The detailed descriptions about the eight datasets are given as follows:

\begin{itemize}

\item{\textit{ETT} \cite{Informer}: This dataset contains the oil temperature and load data collected by electricity transformers, including \textit{ETTh} (\textit{ETTh1} and \textit{ETTh2}) and \textit{ETTm} (\textit{ETTm1} and \textit{ETTm2}), which are sampled hourly and every 15 minutes, respectively.}

\item{\textit{Electricity}\footnote{\url{https://archive.ics.uci.edu/ml/datasets/ElectricityLoadDiagrams20112014}}: This dataset contains the electricity consumption of 321 clients from the UCI Machine Learning Repository, which is sampled hourly.}

\item{\textit{Flight} \cite{MSGNet}: This dataset contains changes in flight data from 7 major European airports provided by OpenSky\footnote{\url{https://opensky-network.org/}}, which is sampled hourly.}


\item{\textit{Weather}\footnote{\url { https://www.bgc-jena.mpg.de/wetter/}}: This dataset contains 21 meteorological measurements data from the Weather Station of the Max Planck Biogeochemistry, which is sampled every 10 minutes.}
\item{\textit{Exchange-Rate}\cite{LSTNet}: This dataset contains the exchange rate data from 8 foreign countries, which is sampled daily.}
\end{itemize}

Following existing works \cite{MSGNet, Informer, Crossformer, pyraformer}, we split each dataset into training, validation, and testing sets based on chronological order. The ratio is 6:2:2 for the \textit{ETT} dataset and 7:2:1 for the others.

\textbf{Experimental Settings.}
MSHyper is implemented in Python with PyTorch 1.13.1 and trained/tested on a single NVIDIA Geforce RTX 3090 GPU, and the source code of MSHyper is released on GitHub \footnote{\url{https://github.com/shangzongjiang/MSHyper}}. We repeat all experiments 3 times and use the mean of the metrics as the final results. Following existing works \cite{MSGNet,TimesNet,DLinear}, we use instance normalization to normalize all datasets. Adam is set as the optimizer. The aggregation windows are set to 4 for \textit{ETT} dataset and 3 for other datasets. $S$ and $H$ in MSHyper are set to 4 in all experiments. For other hyperparameters, we use Neural Network Intelligence (NNI)\footnote{\url{https://nni.readthedocs.io/en/latest/}} toolkit to automatically search the best hyperparameters, which can greatly reduce computation cost compared to the grid search approach. The detailed search space of hyperparameters and the configurations of NNI are given in Table \ref{NNI}.

For the long-range time series forecasting, we have two kinds of settings (i.e., multivariate settings and univariate settings), which are used to forecast all feature dimensions and the last feature dimension, respectively. Following existing works\cite{DLinear, FEDformer}, we use all eight datasets for multivariate long-range time series forecasting and \textit{ETT} dataset for univariate long-range time series forecasting, respectively. 
\begin{table}[]
\centering
\caption{Settings of NNI.\label{NNI}}
\resizebox{0.5\textwidth}{!}{
\begin{tabular}{lllll}
\hline
\textbf{}                     & {Parameters}                     & {Choise}                  \\ \hline
\multirow{4}{*}{Search space} & Batch size                              & \{8, 16, 32, 64, 128\}                               \\
                              & Dropout rate                            &  \{0.05, 0.1, 0.5\}                    \\
                              & Dimension of model                   &  \{64, 128, 256, 512\}                               \\
                              & Initial learning rate          &  \{0.0001, 0.005, 0.01, 0.1\}                               \\
\multirow{3}{*}{Configures}   & Max trial number                        & 120                               \\
                              & Optimization algorithm                  & Tree-structured Parzen Estimator \\ 
                              & Early stopping strategy                 & Curvefitting                      \\ \hline
\end{tabular}
}
\end{table}

\textbf{Evaluation metrics.} Mean Square Error (MSE) and Mean Absolute Error (MAE) are used as evaluation metrics, which are defined as follows:
\begin{equation}
\begin{aligned}
& \mathrm{MSE}=\frac{1}{H} \left\|\widehat{\mathbf X}_{T+1:T+H}^\text{O}-\mathbf X_{T+1:T+H}^\text{O}\right\|^2 \\
& \mathrm{MAE}=\frac{1}{H} \left\|\widehat{\mathbf X}_{T+1:T+H}^\text{O}-\mathbf X_{T+1:T+H}^\text{O}\right\|,
\end{aligned}
\end{equation}
where $T$ and $H$ are the input and output lengths, $\widehat{\mathbf X}_{T+1:T+H}^\text{O}$ and $\mathbf X_{T+1:T+H}^\text{O}$ are forecasting results and ground truth, respectively.
Lower MSE and MAE results mean better performance. 

\subsection{Methods for Comparison}


\textbf{Baselines.} We carefully choose nine well-acknowledged forecasting models as our baselines, including (1) Transformer-based methods: Crossformer \cite{Crossformer}, Pyraformer \cite{pyraformer}, Autoformer \cite{Autoformer}, and Informer\cite{Informer}; (2) Graph-based methods: MSGNet \cite{MSGNet} and MTGNN \cite{MTGNN}; (3) Linear-based methods: DLinear \cite{DLinear} and TiDE \cite{TiDE}; and (4) TCN-based methods: TimesNet \cite{TimesNet}. The detailed descriptions about the nine baselines are as follows:


\begin{itemize}
\item{MSGNet \cite{MSGNet}: It captures inter-scale correlations at different scales by leveraging frequency domain analysis and adaptive graph convolution.}
\item{TimesNet \cite{TimesNet}: It transforms the 1D input sequence into 2D tensors and uses 2D convolution kernels to capture mixed-scale pattern interactions.}
\item{Crossformer \cite{Crossformer}: It combines a two-stage attention with a hierarchical encoder-decoder architecture to capture cross-time and cross-dimension interactions.}
\item{DLinear \cite{DLinear}: It decomposes the input sequence into seasonal and trend components and employs two one-layer linear layers to model each component separately.}
\item{TiDE \cite{TiDE}: It uses dense MLP-based encoder-decoder architectures to handle covariates and non-linear dependencies.}
\item{Pyraformer \cite{pyraformer}: It extends the two-layer structure into multi-scale embeddings, which models the interactions between nodes of different scales through a pyramid graph.}
\item{Autoformer \cite{Autoformer}: It introduces an auto-correlation mechanism to realize seasonal-trend decomposition at the level of sub-sequence.}
\item{Informer \cite{Informer}: It obtains the dominant query by calculating the KL-divergence to reduce the computational complexity.}
\item{MTGNN \cite{MTGNN}: It utilizes a graph learning module to measure inter-variable dependencies and a temporal convolution module to capture temporal pattern interactions.}
\end{itemize}

\subsection{Main Results}

\textbf{Multivariate Results}. Table \ref{tab:table1} shows the multivariate long-range time series forecasting results of MSHyper compared with baselines on all eight datasets, from which we can discern the following tendencies:

\begin{table*}[htbp]
\caption{Multivariate long-range time series forecasting results on eight real-world datasets. The input length is set as $I=96$, and the prediction length $O$ is set as 96, 192, 336, and 720. The best results are \textbf{bolded} and the second best results are \underline{underlined}.}
\label{tab:table1}
\resizebox{\textwidth}{!}{
\begin{tabular}{cc|cc|cc|cc|cc|cc|cc|cc|cc|cc|cc}
\hline
\multicolumn{2}{c|}{Models}                             & \multicolumn{2}{c|}{\begin{tabular}[c]{@{}c@{}}MSHyper\\ (Ours)\end{tabular}} & \multicolumn{2}{c|}{\begin{tabular}[c]{@{}c@{}}MSGNet\\ (2024)\end{tabular}} & \multicolumn{2}{c|}{\begin{tabular}[c]{@{}c@{}}TimesNet\\ (2023)\end{tabular}} & \multicolumn{2}{c|}{\begin{tabular}[c]{@{}c@{}}Crossformer*\\ (2023)\end{tabular}} & \multicolumn{2}{c|}{\begin{tabular}[c]{@{}c@{}}DLinear\\ (2023)\end{tabular}} & \multicolumn{2}{c|}{\begin{tabular}[c]{@{}c@{}}TiDE*\\ (2023)\end{tabular}} & \multicolumn{2}{c|}{\begin{tabular}[c]{@{}c@{}}Pyraformer*\\ (2021)\end{tabular}} & \multicolumn{2}{c|}{\begin{tabular}[c]{@{}c@{}}Autoformer\\ (2021)\end{tabular}} & \multicolumn{2}{c|}{\begin{tabular}[c]{@{}c@{}}Informer\\ (2021)\end{tabular}} & \multicolumn{2}{c}{\begin{tabular}[c]{@{}c@{}}MTGNN\\ (2020)\end{tabular}} \\ \hline
\multicolumn{2}{c|}{Metric}                             & \multicolumn{1}{c}{MSE}                         & MAE                        & \multicolumn{1}{c}{MSE}                        & MAE                        & \multicolumn{1}{c}{MSE}                         & MAE                         & \multicolumn{1}{c}{MSE}                               & MAE                      & \multicolumn{1}{c}{MSE}                             & MAE                    & \multicolumn{1}{c}{MSE}                       & MAE                       & \multicolumn{1}{c}{MSE}                          & MAE                          & \multicolumn{1}{c}{MSE}                          & MAE                          & \multicolumn{1}{c}{MSE}                         & MAE                         & \multicolumn{1}{c}{MSE}                            & MAE                   \\ \hline
\multicolumn{1}{c|}{\multirow{4}{*}{\rotatebox{90}{ETTh1}}}       & 96  & \multicolumn{1}{c}{\textbf{0.383}}              & \textbf{0.392}             & \multicolumn{1}{c}{0.390}                      & 0.411                      & \multicolumn{1}{c}{\underline{0.384}}                       & 0.402                       & \multicolumn{1}{c}{0.418}                             & 0.438                    & \multicolumn{1}{c}{0.386}                           & \underline{0.400}                  & \multicolumn{1}{c}{0.479}                     & 0.464                     & \multicolumn{1}{c}{0.664}                        & 0.510                        & \multicolumn{1}{c}{0.449}                        & 0.459                        & \multicolumn{1}{c}{0.865}                       & 0.713                       & \multicolumn{1}{c}{0.440}                          & 0.450                 \\  
\multicolumn{1}{c|}{}                             & 192 & \multicolumn{1}{c}{\textbf{0.435}}              & \textbf{0.423}             & \multicolumn{1}{c}{0.442}                      & 0.442                      & \multicolumn{1}{c}{\underline{0.436}}                       & \underline{0.429}                       & \multicolumn{1}{c}{0.539}                             & 0.517                    & \multicolumn{1}{c}{0.437}                           & 0.432                  & \multicolumn{1}{c}{0.525}                     & 0.492                     & \multicolumn{1}{c}{0.790}                        & 0.687                        & \multicolumn{1}{c}{0.500}                        & 0.482                        & \multicolumn{1}{c}{1.008}                       & 0.792                       & \multicolumn{1}{c}{0.449}                          & 0.433                 \\ 
\multicolumn{1}{c|}{}                             & 336 & \multicolumn{1}{c}{\textbf{0.480}}              & \textbf{0.445}             & \multicolumn{1}{c}{\textbf{0.480}}                       & 0.468                      & \multicolumn{1}{c}{0.491}                       & 0.469                       & \multicolumn{1}{c}{0.709}                             & 0.638                    & \multicolumn{1}{c}{\underline{0.481}}                           & \underline{0.459}                  & \multicolumn{1}{c}{0.565}                     & 0.515                     & \multicolumn{1}{c}{0.887}                        & 0.738                        & \multicolumn{1}{c}{0.521}                        & 0.496                        & \multicolumn{1}{c}{1.107}                       & 0.809                       & \multicolumn{1}{c}{0.598}                          & 0.554                 \\  
\multicolumn{1}{c|}{}                             & 720 & \multicolumn{1}{c}{\textbf{0.482}}              & \textbf{0.467}             & \multicolumn{1}{c}{\underline{0.494}}                      & \underline{0.488}                      & \multicolumn{1}{c}{0.521}                       & 0.500                       & \multicolumn{1}{c}{0.733}                             & 0.639                    & \multicolumn{1}{c}{0.519}                           & 0.516                  & \multicolumn{1}{c}{0.594}                     & 0.558                     & \multicolumn{1}{c}{0.976}                        & 0.784                        & \multicolumn{1}{c}{0.514}                        & 0.512                        & \multicolumn{1}{c}{1.181}                       & 0.865                       & \multicolumn{1}{c}{0.685}                          & 0.620                 \\ \hline
\multicolumn{1}{c|}{\multirow{4}{*}{\rotatebox{90}{ETTh2}}}       & 96  & \multicolumn{1}{c}{\textbf{0.291}}              & \textbf{0.338}             & \multicolumn{1}{c}{\underline{0.328}}                      & \underline{0.371}                      & \multicolumn{1}{c}{0.340}                       & 0.374                       & \multicolumn{1}{c}{0.425}                             & 0.463                    & \multicolumn{1}{c}{0.333}                           & 0.387                  & \multicolumn{1}{c}{0.400}                     & 0.440                     & \multicolumn{1}{c}{1.392}                        & 0.954                        & \multicolumn{1}{c}{0.346}                        & 0.388                        & \multicolumn{1}{c}{3.755}                       & 1.525                       & \multicolumn{1}{c}{0.496}                          & 0.509                 \\  
\multicolumn{1}{c|}{}                             & 192 & \multicolumn{1}{c}{\textbf{0.376}}              & \textbf{0.391}             & \multicolumn{1}{c}{\underline{0.402}}                      & \underline{0.414}                      & \multicolumn{1}{c}{\underline{0.402}}                       & \underline{0.414}                       & \multicolumn{1}{c}{0.473}                             & 0.500                    & \multicolumn{1}{c}{0.477}                           & 0.476                  & \multicolumn{1}{c}{0.528}                     & 0.509                     & \multicolumn{1}{c}{3.515}                        & 1.561                        & \multicolumn{1}{c}{0.456}                        & 0.452                        & \multicolumn{1}{c}{5.602}                       & 1.931                       & \multicolumn{1}{c}{0.716}                          & 0.616                 \\ 
\multicolumn{1}{c|}{}                             & 336 & \multicolumn{1}{c}{\textbf{0.419}}              & \textbf{0.430}             & \multicolumn{1}{c}{\underline{0.435}}                      & \underline{0.443}                      & \multicolumn{1}{c}{0.452}                       & 0.452                       & \multicolumn{1}{c}{0.581}                             & 0.562                    & \multicolumn{1}{c}{0.594}                           & 0.541                  & \multicolumn{1}{c}{0.643}                     & 0.571                     & \multicolumn{1}{c}{4.471}                        & 1.835                        & \multicolumn{1}{c}{0.482}                        & 0.486                        & \multicolumn{1}{c}{4.721}                       & 1.835                       & \multicolumn{1}{c}{0.718}                          & 0.614                 \\  
\multicolumn{1}{c|}{}                             & 720 & \multicolumn{1}{c}{\textbf{0.417}}                       & \textbf{0.435}                      & \multicolumn{1}{c}{\textbf{0.417}}             & \underline{0.441}             & \multicolumn{1}{c}{\underline{0.462}}                       & 0.468                       & \multicolumn{1}{c}{0.775}                             & 0.665                    & \multicolumn{1}{c}{0.831}                           & 0.657                  & \multicolumn{1}{c}{0.874}                     & 0.679                     & \multicolumn{1}{c}{4.190}                        & 1.770                        & \multicolumn{1}{c}{0.515}                        & 0.511                        & \multicolumn{1}{c}{3.647}                       & 1.625                       & \multicolumn{1}{c}{1.161}                          & 0.791                 \\ \hline
\multicolumn{1}{c|}{\multirow{4}{*}{\rotatebox{90}{ETTm1}}}       & 96  & \multicolumn{1}{c}{\underline{0.331}}                       & \textbf{0.350}                      & \multicolumn{1}{c}{\textbf{0.319}}             & \underline{0.366}             & \multicolumn{1}{c}{{0.338}}                       & 0.375                       & \multicolumn{1}{c}{0.361}                             & 0.403                    & \multicolumn{1}{c}{0.345}                           & 0.372                  & \multicolumn{1}{c}{0.364}                     & 0.387                     & \multicolumn{1}{c}{0.535}                        & 0.510                        & \multicolumn{1}{c}{0.505}                        & 0.475                        & \multicolumn{1}{c}{0.672}                       & 0.571                       & \multicolumn{1}{c}{0.381}                          & 0.415                 \\ 
\multicolumn{1}{c|}{}                             & 192 & \multicolumn{1}{c}{\textbf{0.374}}                       & \textbf{0.373}             & \multicolumn{1}{c}{\underline{0.376}}                      & 0.397                      & \multicolumn{1}{c}{\textbf{0.374}}              & \underline{0.387}              & \multicolumn{1}{c}{0.387}                             & 0.422                    & \multicolumn{1}{c}{0.380}                           & 0.389                  & \multicolumn{1}{c}{0.398}                     & 0.404                     & \multicolumn{1}{c}{0.580}                        & 0.549                        & \multicolumn{1}{c}{0.553}                        & 0.496                        & \multicolumn{1}{c}{0.795}                       & 0.669                       & \multicolumn{1}{c}{0.442}                          & 0.451                 \\  
\multicolumn{1}{c|}{}                             & 336 & \multicolumn{1}{c}{\textbf{0.408}}                       & \textbf{0.395}             & \multicolumn{1}{c}{0.417}                      & 0.422                      & \multicolumn{1}{c}{\underline{0.410}}              & \underline{0.411}                       & \multicolumn{1}{c}{0.605}                             & 0.572                    & \multicolumn{1}{c}{{0.413}}                           & 0.413                  & \multicolumn{1}{c}{0.428}                     & 0.425                     & \multicolumn{1}{c}{0.736}                        & 0.646                        & \multicolumn{1}{c}{0.621}                        & 0.537                        & \multicolumn{1}{c}{1.212}                       & 0.871                       & \multicolumn{1}{c}{0.475}                          & 0.475                 \\  
\multicolumn{1}{c|}{}                             & 720 & \multicolumn{1}{c}{\textbf{0.473}}                       & \textbf{0.431}             & \multicolumn{1}{c}{0.481}                      & 0.458                      & \multicolumn{1}{c}{\textbf{0.473}}                       & \underline{0.450}                       & \multicolumn{1}{c}{0.703}                             & 0.645                    & \multicolumn{1}{c}{\underline{0.474}}                  & 0.453                  & \multicolumn{1}{c}{0.487}                     & 0.461                     & \multicolumn{1}{c}{1.056}                        & 0.782                        & \multicolumn{1}{c}{0.671}                        & 0.561                        & \multicolumn{1}{c}{1.166}                       & 0.823                       & \multicolumn{1}{c}{0.531}                          & 0.507                 \\ \hline
\multicolumn{1}{c|}{\multirow{4}{*}{\rotatebox{90}{ETTm2}}}       & 96  & \multicolumn{1}{c}{\underline{0.179}}                       & {\textbf{0.257}}                      & \multicolumn{1}{c}{\textbf{0.177}}             & \underline{0.262}             & \multicolumn{1}{c}{0.187}                       & 0.267                       & \multicolumn{1}{c}{0.275}                             & 0.358                    & \multicolumn{1}{c}{0.193}                           & 0.292                  & \multicolumn{1}{c}{0.207}                     & 0.305                     & \multicolumn{1}{c}{0.361}                        & 0.450                        & \multicolumn{1}{c}{0.255}                        & 0.339                        & \multicolumn{1}{c}{0.365}                       & 0.453                       & \multicolumn{1}{c}{0.240}                          & 0.343                 \\  
\multicolumn{1}{c|}{}                             & 192 & \multicolumn{1}{c}{\textbf{0.247}}              & \textbf{0.305}             & \multicolumn{1}{c}{\textbf{0.247}}             & \underline{0.307}                      & \multicolumn{1}{c}{\underline{0.249}}                       & 0.309                       & \multicolumn{1}{c}{0.345}                             & 0.400                    & \multicolumn{1}{c}{0.284}                           & 0.362                  & \multicolumn{1}{c}{0.290}                     & 0.364                     & \multicolumn{1}{c}{0.720}                        & 0.667                        & \multicolumn{1}{c}{0.281}                        & 0.340                        & \multicolumn{1}{c}{0.533}                       & 0.563                       & \multicolumn{1}{c}{0.398}                          & 0.454                 \\ 
\multicolumn{1}{c|}{}                             & 336 & \multicolumn{1}{c}{\textbf{0.309}}              & \textbf{0.344}             & \multicolumn{1}{c}{\underline{0.312}}                      & \underline{0.346}                      & \multicolumn{1}{c}{0.321}                       & 0.351                       & \multicolumn{1}{c}{0.657}                             & 0.528                    & \multicolumn{1}{c}{0.369}                           & 0.427                  & \multicolumn{1}{c}{0.377}                     & 0.422                     & \multicolumn{1}{c}{1.294}                        & 0.882                        & \multicolumn{1}{c}{0.339}                        & 0.372                        & \multicolumn{1}{c}{1.363}                       & 0.887                       & \multicolumn{1}{c}{0.568}                          & 0.555                 \\  
\multicolumn{1}{c|}{}                             & 720 & \multicolumn{1}{c}{\textbf{0.401}}              & \textbf{0.398}             & \multicolumn{1}{c}{0.414}                      & \underline{0.403}                      & \multicolumn{1}{c}{\underline{0.408}}                       & \underline{0.403}                       & \multicolumn{1}{c}{1.208}                             & 0.753                    & \multicolumn{1}{c}{0.554}                           & 0.522                  & \multicolumn{1}{c}{0.558}                     & 0.524                     & \multicolumn{1}{c}{4.147}                        & 1.547                        & \multicolumn{1}{c}{0.433}                        & 0.432                        & \multicolumn{1}{c}{3.379}                       & 1.338                       & \multicolumn{1}{c}{1.072}                          & 0.767                 \\ \hline
\multicolumn{1}{c|}{\multirow{4}{*}{\rotatebox{90}{Flight}}}      & 96  & \multicolumn{1}{c}{\textbf{0.155}}              & \textbf{0.260}             & \multicolumn{1}{c}{\underline{0.183}}                      & \underline{0.301}                      & \multicolumn{1}{c}{0.237}                       & 0.350                       & \multicolumn{1}{c}{0.410}                             & 0.449                    & \multicolumn{1}{c}{0.221}                           & 0.337                  & \multicolumn{1}{c}{0.247}                     & 0.440                     & \multicolumn{1}{c}{0.452}                        & 0.508                        & \multicolumn{1}{c}{0.204}                        & 0.319                        & \multicolumn{1}{c}{0.333}                       & 0.405                       & \multicolumn{1}{c}{0.196}                          & 0.316                 \\ 
\multicolumn{1}{c|}{}                             & 192 & \multicolumn{1}{c}{\textbf{0.155}}              & \textbf{0.258}             & \multicolumn{1}{c}{\underline{0.189}}                      & \underline{0.306}                      & \multicolumn{1}{c}{0.224}                       & 0.337                       & \multicolumn{1}{c}{0.503}                             & 0.512                    & \multicolumn{1}{c}{0.220}                           & 0.336                  & \multicolumn{1}{c}{0.303}                     & 0.493                     & \multicolumn{1}{c}{0.458}                        & 0.507                        & \multicolumn{1}{c}{0.200}                        & 0.314                        & \multicolumn{1}{c}{0.358}                       & 0.421                       & \multicolumn{1}{c}{0.272}                          & 0.379                 \\ 
\multicolumn{1}{c|}{}                             & 336 & \multicolumn{1}{c}{\textbf{0.164}}              & \textbf{0.266}             & \multicolumn{1}{c}{\underline{0.206}}                      & \underline{0.320}                      & \multicolumn{1}{c}{0.289}                       & 0.394                       & \multicolumn{1}{c}{0.544}                             & 0.532                    & \multicolumn{1}{c}{0.229}                           & 0.342                  & \multicolumn{1}{c}{0.354}                     & 0.533                     & \multicolumn{1}{c}{0.483}                        & 0.522                        & \multicolumn{1}{c}{0.201}                        & 0.318                        & \multicolumn{1}{c}{0.398}                       & 0.446                       & \multicolumn{1}{c}{0.260}                          & 0.369                 \\  
\multicolumn{1}{c|}{}                             & 720 & \multicolumn{1}{c}{\textbf{0.202}}              & \textbf{0.299}             & \multicolumn{1}{c}{\underline{0.253}}                      & \underline{0.358}                      & \multicolumn{1}{c}{0.310}                       & 0.408                       & \multicolumn{1}{c}{0.702}                             & 0.635                    & \multicolumn{1}{c}{0.263}                           & 0.366                  & \multicolumn{1}{c}{0.572}                     & 0.689                     & \multicolumn{1}{c}{0.521}                        & 0.536                        & \multicolumn{1}{c}{0.345}                        & 0.426                        & \multicolumn{1}{c}{0.476}                       & 0.484                       & \multicolumn{1}{c}{0.390}                          & 0.449                 \\ \hline
\multicolumn{1}{c|}{\multirow{4}{*}{\rotatebox{90}{Weather}}}     & 96  & \multicolumn{1}{c}{\textbf{0.157}}                       & \textbf{0.198}             & \multicolumn{1}{c}{\underline{0.163}}                      & \underline{0.212}                      & \multicolumn{1}{c}{0.172}                       & 0.220                       & \multicolumn{1}{c}{{0.232}}                    & 0.302                    & \multicolumn{1}{c}{0.196}                           & 0.255                  & \multicolumn{1}{c}{0.202}                     & 0.261                     & \multicolumn{1}{c}{0.211}                        & 0.295                        & \multicolumn{1}{c}{0.266}                        & 0.336                        & \multicolumn{1}{c}{0.300}                       & 0.384                       & \multicolumn{1}{c}{0.171}                          & 0.231                 \\  
\multicolumn{1}{c|}{}                             & 192 & \multicolumn{1}{c}{\textbf{0.207}}                       & \textbf{0.244}             & \multicolumn{1}{c}{\underline{0.212}}                      & \underline{0.254}                      & \multicolumn{1}{c}{0.219}                       & 0.261                       & \multicolumn{1}{c}{{0.371}}                    & 0.410                    & \multicolumn{1}{c}{0.237}                           & 0.296                  & \multicolumn{1}{c}{0.242}                     & 0.298                     & \multicolumn{1}{c}{0.240}                        & 0.316                        & \multicolumn{1}{c}{0.307}                        & 0.367                        & \multicolumn{1}{c}{0.598}                       & 0.544                       & \multicolumn{1}{c}{0.215}                          & 0.274                 \\ 
\multicolumn{1}{c|}{}                             & 336 & \multicolumn{1}{c}{\textbf{0.265}}                       & \textbf{0.286}             & \multicolumn{1}{c}{{0.272}}                      & \underline{0.299}                      & \multicolumn{1}{c}{0.280}                       & 0.306                       & \multicolumn{1}{c}{0.495}                             & 0.515                    & \multicolumn{1}{c}{0.283}                           & 0.335                  & \multicolumn{1}{c}{0.287}                     & 0.335                     & \multicolumn{1}{c}{0.295}                        & 0.356                        & \multicolumn{1}{c}{0.359}                        & 0.395                        & \multicolumn{1}{c}{0.578}                       & 0.523                       & \multicolumn{1}{c}{\underline{0.266}}                 & 0.313                 \\ 
\multicolumn{1}{c|}{}                             & 720 & \multicolumn{1}{c}{\textbf{0.342}}                       & \textbf{0.339}             & \multicolumn{1}{c}{{0.350}}                      & \underline{0.348}                      & \multicolumn{1}{c}{0.365}                       & 0.359                       & \multicolumn{1}{c}{0.526}                             & 0.542                    & \multicolumn{1}{c}{{0.345}}                  & 0.381                  & \multicolumn{1}{c}{0.351}                     & 0.386                     & \multicolumn{1}{c}{0.368}                        & 0.407                        & \multicolumn{1}{c}{0.419}                        & 0.428                        & \multicolumn{1}{c}{1.059}                       & 0.741                       & \multicolumn{1}{c}{{\underline{0.344}}}                 & 0.375                 \\ \hline
\multicolumn{1}{c|}{\multirow{4}{*}{\rotatebox{90}{Exchange}}}    & 96  & \multicolumn{1}{c}{\textbf{0.083}}              & \textbf{0.199}             & \multicolumn{1}{c}{0.102}                      & 0.230                      & \multicolumn{1}{c}{0.107}                       & 0.234                       & \multicolumn{1}{c}{0.256}                             & 0.367                    & \multicolumn{1}{c}{\underline{0.088}}                           & \underline{0.218}                  & \multicolumn{1}{c}{0.094}                     & 0.218                     & \multicolumn{1}{c}{1.468}                        & 1.002                        & \multicolumn{1}{c}{0.197}                        & 0.323                        & \multicolumn{1}{c}{0.847}                       & 0.752                       & \multicolumn{1}{c}{0.267}                          & 0.378                 \\ 
\multicolumn{1}{c|}{}                             & 192 & \multicolumn{1}{c}{\textbf{0.173}}              & \textbf{0.294}             & \multicolumn{1}{c}{0.195}                      & 0.317                      & \multicolumn{1}{c}{0.226}                       & 0.344                       & \multicolumn{1}{c}{0.470}                             & 0.509                    & \multicolumn{1}{c}{\underline{0.176}}                           & \underline{0.315}                  & \multicolumn{1}{c}{0.184}                     & 0.307                     & \multicolumn{1}{c}{1.583}                        & 1.061                        & \multicolumn{1}{c}{0.300}                        & 0.369                        & \multicolumn{1}{c}{1.204}                       & 0.895                       & \multicolumn{1}{c}{0.590}                          & 0.578                 \\  
\multicolumn{1}{c|}{}                             & 336 & \multicolumn{1}{c}{\textbf{0.310}}                       & \textbf{0.411}             & \multicolumn{1}{c}{0.359}                      & 0.436                      & \multicolumn{1}{c}{0.367}                       & 0.448                       & \multicolumn{1}{c}{1.268}                             & 0.883                    & \multicolumn{1}{c}{\underline{0.313}}                  & \underline{0.427}                  & \multicolumn{1}{c}{0.349}                     & 0.431                     & \multicolumn{1}{c}{1.733}                        & 1.110                        & \multicolumn{1}{c}{0.509}                        & 0.524                        & \multicolumn{1}{c}{1.672}                       & 1.036                       & \multicolumn{1}{c}{0.939}                          & 0.749                 \\ 
\multicolumn{1}{c|}{}                             & 720 & \multicolumn{1}{c}{\underline{0.846}}                       & \textbf{0.692}             & \multicolumn{1}{c}{0.940}                      & 0.738                      & \multicolumn{1}{c}{0.964}                       & 0.746                       & \multicolumn{1}{c}{1.767}                             & 1.068                    & \multicolumn{1}{c}{\textbf{0.839}}                  & \underline{0.695}                  & \multicolumn{1}{c}{0.852}                     & 0.698                     & \multicolumn{1}{c}{2.000}                        & 1.184                        & \multicolumn{1}{c}{1.447}                        & 0.941                        & \multicolumn{1}{c}{2.478}                       & 1.310                       & \multicolumn{1}{c}{1.107}                          & 0.834                 \\ \hline
\multicolumn{1}{c|}{\multirow{4}{*}{\rotatebox{90}{Electricity}}} & 96  & \multicolumn{1}{c}{\textbf{0.152}}              & \textbf{0.252}             & \multicolumn{1}{c}{\underline{0.165}}                      & 0.274                      & \multicolumn{1}{c}{0.168}                       & \underline{0.272}                       & \multicolumn{1}{c}{0.219}                             & 0.314                    & \multicolumn{1}{c}{0.197}                           & 0.282                  & \multicolumn{1}{c}{0.237}                     & 0.329                     & \multicolumn{1}{c}{0.372}                        & 0.441                        & \multicolumn{1}{c}{0.201}                        & 0.317                        & \multicolumn{1}{c}{0.274}                       & 0.368                       & \multicolumn{1}{c}{0.211}                          & 0.305                 \\  
\multicolumn{1}{c|}{}                             & 192 & \multicolumn{1}{c}{\textbf{0.171}}              & \textbf{0.271}             & \multicolumn{1}{c}{\underline{0.184}}                      & 0.292                      & \multicolumn{1}{c}{\underline{0.184}}                       & 0.289                       & \multicolumn{1}{c}{0.231}                             & 0.322                    & \multicolumn{1}{c}{0.196}                           & \underline{0.285}                  & \multicolumn{1}{c}{0.236}                     & 0.330                     & \multicolumn{1}{c}{0.370}                        & 0.440                        & \multicolumn{1}{c}{0.222}                        & 0.334                        & \multicolumn{1}{c}{0.296}                       & 0.386                       & \multicolumn{1}{c}{0.225}                          & 0.319                 \\  
\multicolumn{1}{c|}{}                             & 336 & \multicolumn{1}{c}{\textbf{0.187}}              & \textbf{0.284}             & \multicolumn{1}{c}{\underline{0.195}}                      & 0.302                      & \multicolumn{1}{c}{0.198}                       & \underline{0.300}                       & \multicolumn{1}{c}{0.246}                             & 0.337                    & \multicolumn{1}{c}{0.209}                           & 0.301                  & \multicolumn{1}{c}{0.249}                     & 0.344                     & \multicolumn{1}{c}{0.368}                        & 0.440                        & \multicolumn{1}{c}{0.231}                        & 0.338                        & \multicolumn{1}{c}{0.300}                       & 0.394                       & \multicolumn{1}{c}{0.247}                          & 0.340                 \\  
\multicolumn{1}{c|}{}                             & 720 & \multicolumn{1}{c}{\underline{0.224}}                       & \textbf{0.316}             & \multicolumn{1}{c}{0.231}                      & 0.332                      & \multicolumn{1}{c}{\textbf{0.220}}              & \underline{0.320}                       & \multicolumn{1}{c}{0.280}                             & 0.363                    & \multicolumn{1}{c}{0.245}                           & 0.333                  & \multicolumn{1}{c}{0.284}                     & 0.373                     & \multicolumn{1}{c}{0.373}                        & 0.446                        & \multicolumn{1}{c}{0.254}                        & 0.361                        & \multicolumn{1}{c}{0.373}                       & 0.439                       & \multicolumn{1}{c}{0.287}                          & 0.373                 \\ \hline
\end{tabular}

}
\begin{tablenotes}
\item{* indicates that some methods do not have uniform prediction lengths with other methods. To ensure a fair comparison, we utilize their official code and adjust prediction lengths. Other results are from MSGNet.}
\end{tablenotes}
\end{table*}

\begin{itemize}
\item {MSHyper gets the SOTA results on all eight datasets and even achieves better performance on the datasets with low auto-correlation function (ACF) values (e.g., \textit{Exchange-Rate} and \textit{ETTm1} datasets).
Specifically, MSHyper reduces prediction errors by an average of 4.06\% and 5.27\% over the best baseline in MSE and MAE, respectively.
We attribute this to the reason that MSHyper can get multi-scale embeddings and capture high-order interactions between temporal patterns of different scales. In addition, prediction errors increase smoothly and slowly when the forecasting horizon is increased, which means that the MSHyper retains better long-range robustness. These observations provide empirical guidance for the success of using MSHyper in multivariate long-range time series forecasting.} 


\item {Traditional transformer-based methods (i.e., Informer, Autoformer, and Pyraformer) exhibit relatively poor predictive performance. This is because time series forecasting requires modeling temporal pattern interactions, while vanilla attention or simplistic decomposition techniques are insufficient in capturing such interactions.}

\item {Although using different backbones, existing multi-scale decomposition methods (i.e., MTGNN, DLinear, TiDE, Crossformer, TimesNet, and MSGNet) all achieve competitive performance. Specially, by considering multi-scale pattern interactions on the decomposed sequences, MSGNet and TimesNet achieve promising forecasting results. However, they fail to consider high-order pattern interactions and get worse performance than MSHyper in most cases.}
\end{itemize}

 \textbf{Univariate Results.} Table \ref{tab:table2} summarizes the univariate long-range time series forecasting results of MSHyper compared with baselines on \textit{ETT} dataset, from which we can discern the following tendencies:

\begin{itemize}
\item {MSHyper still achieves the SOTA performance for univariate long-range time series forecasting, and the prediction errors increase steadily and slowly when increasing the forecasting horizon, which demonstrates the effectiveness of MSHyper in improving the capability of univariate long-range time series forecasting.}

\item { Although MLP-based methods perform better than traditional Transformer-based methods, they fail to capture multi-scale pattern interactions and get worse performance than MSGNet, TimesNet, and MSHyper.}
\item { MSGNet and TimesNet are the best baselines that use multi-head attention mechanism and 2D convolution kernels to learn multi-scale pattern interactions, respectively. However, they only consider one type of multi-scale (i.e., intra-scale or mixed-scale) pattern interactions and get worse performance than MSHyper. In contrast, MSHyper leverages multi-scale hypergraph structures to capture intra-scale, inter-scale, and mixed-scale pattern interactions.}
\end{itemize}

\begin{table*}[]
\centering
\caption{Univariate long-range time series forecasting results on \textit{ETT} dataset. The input length is set as $I$=96, and the prediction length $O$ is set as 96, 192, 336, and 720. The best results are \textbf{bolded} and the second best results are \underline{underlined}.}
\resizebox{\textwidth}{!}{
\label{tab:table2}
\begin{tabular}{cc|cc|cc|cc|cc|cc|cc|cc|cc|cc|cc}
\hline
\multicolumn{2}{c|}{Models}                       & \multicolumn{2}{c|}{\begin{tabular}[c]{@{}c@{}}MSHyper\\ (Ours)\end{tabular}} & \multicolumn{2}{c|}{\begin{tabular}[c]{@{}c@{}}MSGNet\\ (2024)\end{tabular}} & \multicolumn{2}{c|}{\begin{tabular}[c]{@{}c@{}}TimesNet\\ (2023)\end{tabular}} & \multicolumn{2}{c|}{\begin{tabular}[c]{@{}c@{}}Crossformer\\ (2023)\end{tabular}} & \multicolumn{2}{c|}{\begin{tabular}[c]{@{}c@{}}DLinear\\ (2023)\end{tabular}} & \multicolumn{2}{c|}{\begin{tabular}[c]{@{}c@{}}TiDE\\ (2023)\end{tabular}} & \multicolumn{2}{c|}{\begin{tabular}[c]{@{}c@{}}Pyraformer\\ (2021)\end{tabular}} & \multicolumn{2}{c|}{\begin{tabular}[c]{@{}c@{}}Autoformer\\ (2021)\end{tabular}} & \multicolumn{2}{c|}{\begin{tabular}[c]{@{}c@{}}Informer\\ (2021)\end{tabular}} & \multicolumn{2}{c}{\begin{tabular}[c]{@{}c@{}}MTGNN\\ (2020)\end{tabular}} \\ \hline
\multicolumn{2}{c|}{Metric}                       & \multicolumn{1}{c}{MSE}                         & MAE                        & \multicolumn{1}{c}{MSE}                         & MAE                       & \multicolumn{1}{c}{MSE}                         & MAE                         & \multicolumn{1}{c}{MSE}                               & MAE                      & \multicolumn{1}{c}{MSE}                         & MAE                        & \multicolumn{1}{c}{MSE}                       & MAE                       & \multicolumn{1}{c}{MSE}                          & MAE                          & \multicolumn{1}{c}{MSE}                          & MAE                          & \multicolumn{1}{c}{MSE}                         & MAE                         & \multicolumn{1}{c}{MSE}                        & MAE                       \\ \hline
\multicolumn{1}{c|}{\multirow{4}{*}{\rotatebox{90}{ETTh1}}} & 96  & \multicolumn{1}{c}{\textbf{0.055}}              & \textbf{0.178}             & \multicolumn{1}{c}{{0.059}}                       & {0.186}                     & \multicolumn{1}{c}{\underline{0.058}}                       & {0.185}                       & \multicolumn{1}{c}{0.076}                             & {0.216}                    & \multicolumn{1}{c}{0.061}                       & \underline{0.184}                      & \multicolumn{1}{c}{0.068}                     & 0.193                     & \multicolumn{1}{c}{0.461}                        & 0.621                        & \multicolumn{1}{c}{0.449}                        & 0.459                        & \multicolumn{1}{c}{0.865}                       & 0.713                       & \multicolumn{1}{c}{0.382}                      & 0.560                      \\  
\multicolumn{1}{c|}{}                       & 192 & \multicolumn{1}{c}{\textbf{0.076}}                       & \textbf{0.209}             & \multicolumn{1}{c}{{0.078}}              & 0.214                     & \multicolumn{1}{c}{\underline{0.077}}                       & \underline{0.213}                       & \multicolumn{1}{c}{{0.085}}                    & {0.225}                    & \multicolumn{1}{c}{0.083}                       & {0.214}             & \multicolumn{1}{c}{0.083}                     & 0.216                     & \multicolumn{1}{c}{0.620}                         & 0.731                        & \multicolumn{1}{c}{0.500}                          & 0.482                        & \multicolumn{1}{c}{1.008}                       & 0.792                       & \multicolumn{1}{c}{0.470}                       & 0.624                     \\ 
\multicolumn{1}{c|}{}                       & 336 & \multicolumn{1}{c}{\textbf{0.090}}                        & \textbf{0.237}                      & \multicolumn{1}{c}{\underline{0.096}}                       & \underline{0.242}                     & \multicolumn{1}{c}{{0.104}}              & {0.249}              & \multicolumn{1}{c}{{0.106}}                             & {0.257}                    & \multicolumn{1}{c}{0.100}                         & 0.245                      & \multicolumn{1}{c}{0.098}                     & 0.244                     & \multicolumn{1}{c}{0.494}                        & 0.650                         & \multicolumn{1}{c}{0.521}                        & 0.496                        & \multicolumn{1}{c}{1.107}                       & 0.809                       & \multicolumn{1}{c}{0.381}                      & 0.553                     \\  
\multicolumn{1}{c|}{}                       & 720 & \multicolumn{1}{c}{\underline{0.098}}                       & \underline{0.246}                      & \multicolumn{1}{c}{0.126}                       & 0.278                     & \multicolumn{1}{c}{\textbf{0.086}}              & \textbf{0.232}              & \multicolumn{1}{c}{0.128}                             & 0.287                    & \multicolumn{1}{c}{0.197}                       & 0.369                      & \multicolumn{1}{c}{0.178}                     & 0.347                     & \multicolumn{1}{c}{0.697}                        & 0.784                        & \multicolumn{1}{c}{0.514}                        & 0.512                        & \multicolumn{1}{c}{1.181}                       & 0.865                       & \multicolumn{1}{c}{0.313}                      & 0.498                     \\ \hline
\multicolumn{1}{c|}{\multirow{4}{*}{\rotatebox{90}{ETTh2}}} & 96  & \multicolumn{1}{c}{\textbf{0.127}}              & \underline{0.274}                      & \multicolumn{1}{c}{0.144}                       & 0.295                     & \multicolumn{1}{c}{{0.141}}                       & 0.293                       & \multicolumn{1}{c}{0.138}                             & 0.289                    & \multicolumn{1}{c}{\underline{0.128}}              & \textbf{0.271}             & \multicolumn{1}{c}{0.145}                     & 0.296                     & \multicolumn{1}{c}{0.405}                        & 0.519                        & \multicolumn{1}{c}{0.358}                        & 0.397                        & \multicolumn{1}{c}{3.755}                       & 1.525                       & \multicolumn{1}{c}{0.516}                      & 0.620                      \\  
\multicolumn{1}{c|}{}                       & 192 & \multicolumn{1}{c}{\textbf{0.178}}              & \textbf{0.327}                      & \multicolumn{1}{c}{0.198}                       & 0.353                     & \multicolumn{1}{c}{0.192}                       & 0.345                       & \multicolumn{1}{c}{0.188}                             & 0.341                    & \multicolumn{1}{c}{{0.182}}                        & \underline{0.328}             & \multicolumn{1}{c}{\underline{0.179}}                     & 0.346                     & \multicolumn{1}{c}{1.528}                        & 1.102                        & \multicolumn{1}{c}{0.456}                        & 0.452                        & \multicolumn{1}{c}{5.602}                       & 1.931                       & \multicolumn{1}{c}{1.021}                      & 0.902                     \\  
\multicolumn{1}{c|}{}                       & 336 & \multicolumn{1}{c}{\textbf{0.217}}              & \underline{0.372}             & \multicolumn{1}{c}{\underline{0.224}}                       & 0.381                     & \multicolumn{1}{c}{0.240}                       & 0.394                       & \multicolumn{1}{c}{0.229}                             & 0.384                    & \multicolumn{1}{c}{0.231}                       & {0.377}                      & \multicolumn{1}{c}{0.226}                     & \textbf{0.367}                     & \multicolumn{1}{c}{1.546}                        & 1.110                         & \multicolumn{1}{c}{0.482}                        & 0.486                        & \multicolumn{1}{c}{4.721}                       & 1.835                       & \multicolumn{1}{c}{1.496}                      & 1.108                     \\ 
\multicolumn{1}{c|}{}                       & 720 & \multicolumn{1}{c}{\underline{0.228}}                       & \underline{0.383}                      & \multicolumn{1}{c}{\textbf{0.222}}              & \textbf{0.380}             & \multicolumn{1}{c}{0.238}                       & 0.390                       & \multicolumn{1}{c}{0.250}                             & 0.404                    & \multicolumn{1}{c}{0.322}                       & 0.464                      & \multicolumn{1}{c}{0.319}                     & 0.462                     & \multicolumn{1}{c}{1.302}                        & 1.016                        & \multicolumn{1}{c}{0.515}                        & 0.511                        & \multicolumn{1}{c}{3.647}                       & 1.625                       & \multicolumn{1}{c}{1.969}                      & 1.306                     \\ \hline
\multicolumn{1}{c|}{\multirow{4}{*}{\rotatebox{90}{ETTm1}}} & 96  & \multicolumn{1}{c}{\textbf{0.028}}              & \textbf{0.125}             & \multicolumn{1}{c}{\textbf{0.028}}              & \underline{0.126}                     & \multicolumn{1}{c}{\underline{0.029}}                       & 0.127                       & \multicolumn{1}{c}{0.071}                             & 0.193                    & \multicolumn{1}{c}{0.034}                       & 0.135                      & \multicolumn{1}{c}{0.034}                     & 0.136                     & \multicolumn{1}{c}{0.089}                        & 0.239                        & \multicolumn{1}{c}{0.505}                        & 0.475                        & \multicolumn{1}{c}{0.672}                       & 0.571                       & \multicolumn{1}{c}{0.112}                      & 0.271                     \\  
\multicolumn{1}{c|}{}                       & 192 & \multicolumn{1}{c}{\textbf{0.042}}              & \textbf{0.156}             & \multicolumn{1}{c}{\underline{0.044}}                       & \underline{0.160}                      & \multicolumn{1}{c}{0.047}                       & 0.163                       & \multicolumn{1}{c}{0.106}                             & 0.245                    & \multicolumn{1}{c}{0.054}                       & 0.174                      & \multicolumn{1}{c}{0.055}                     & 0.174                     & \multicolumn{1}{c}{0.214}                        & 0.388                        & \multicolumn{1}{c}{0.553}                        & 0.496                        & \multicolumn{1}{c}{0.795}                       & 0.669                       & \multicolumn{1}{c}{0.217}                      & 0.393                     \\  
\multicolumn{1}{c|}{}                       & 336 & \multicolumn{1}{c}{\textbf{0.056}}              & \textbf{0.181}             & \multicolumn{1}{c}{0.059}                       & \underline{0.187}                     & \multicolumn{1}{c}{\underline{0.058}}                       & \underline{0.187}                       & \multicolumn{1}{c}{0.104}                             & 0.287                    & \multicolumn{1}{c}{0.075}                       & 0.204                      & \multicolumn{1}{c}{0.073}                     & 0.202                     & \multicolumn{1}{c}{0.332}                        & 0.505                        & \multicolumn{1}{c}{0.621}                        & 0.537                        & \multicolumn{1}{c}{1.212}                       & 0.871                       & \multicolumn{1}{c}{0.389}                      & 0.556                     \\  
\multicolumn{1}{c|}{}                       & 720 & \multicolumn{1}{c}{\textbf{0.078}}              & \textbf{0.215}             & \multicolumn{1}{c}{\underline{0.081}}                       & \underline{0.218}                     & \multicolumn{1}{c}{\underline{0.081}}                       & 0.220                       & \multicolumn{1}{c}{0.183}                             & 0.334                    & \multicolumn{1}{c}{0.113}                       & 0.254                      & \multicolumn{1}{c}{0.100}                     & 0.236                     & \multicolumn{1}{c}{0.543}                        & 0.662                        & \multicolumn{1}{c}{0.671}                        & 0.561                        & \multicolumn{1}{c}{1.166}                       & 0.823                       & \multicolumn{1}{c}{0.633}                      & 0.719                     \\ \hline
\multicolumn{1}{c|}{\multirow{4}{*}{\rotatebox{90}{ETTm2}}} & 96  & \multicolumn{1}{c}{\textbf{0.069}}              & \textbf{0.189}             & \multicolumn{1}{c}{0.074}                       & 0.206                     & \multicolumn{1}{c}{0.075}                       & 0.201                       & \multicolumn{1}{c}{0.071}                             & 0.194                    & \multicolumn{1}{c}{\underline{0.070}}                        & \underline{0.191}                      & \multicolumn{1}{c}{0.073}                     & 0.197                     & \multicolumn{1}{c}{0.167}                        & 0.314                        & \multicolumn{1}{c}{0.255}                        & 0.339                        & \multicolumn{1}{c}{0.365}                       & 0.453                       & \multicolumn{1}{c}{0.156}                      & 0.313                     \\ 
\multicolumn{1}{c|}{}                       & 192 & \multicolumn{1}{c}{\textbf{0.099}}              & \textbf{0.235}             & \multicolumn{1}{c}{0.108}                       & 0.249                     & \multicolumn{1}{c}{0.107}                       & 0.248                       & \multicolumn{1}{c}{0.105}                             & 0.242                    & \multicolumn{1}{c}{\underline{0.104}}                       & \underline{0.238}                      & \multicolumn{1}{c}{0.107}                     & 0.242                     & \multicolumn{1}{c}{0.550}                         & 0.603                        & \multicolumn{1}{c}{0.281}                        & 0.340                        & \multicolumn{1}{c}{0.533}                       & 0.563                       & \multicolumn{1}{c}{0.194}                      & 0.341                     \\ 
\multicolumn{1}{c|}{}                       & 336 & \multicolumn{1}{c}{\textbf{0.127}}              & \textbf{0.270}             & \multicolumn{1}{c}{0.140}                       & 0.287                     & \multicolumn{1}{c}{0.142}                       & 0.289                       & \multicolumn{1}{c}{0.135}                             & 0.281                    & \multicolumn{1}{c}{\underline{0.134}}                       & \underline{0.277}                      & \multicolumn{1}{c}{0.138}                     & 0.282                     & \multicolumn{1}{c}{0.873}                        & 0.757                        & \multicolumn{1}{c}{0.339}                        & 0.372                        & \multicolumn{1}{c}{1.363}                       & 0.887                       & \multicolumn{1}{c}{0.816}                      & 0.809                     \\  
\multicolumn{1}{c|}{}                       & 720 & \multicolumn{1}{c}{\textbf{0.176}}              & \textbf{0.326}             & \multicolumn{1}{c}{0.194}                       & 0.343                     & \multicolumn{1}{c}{0.206}                       & 0.352                       & \multicolumn{1}{c}{0.189}                             & 0.340                    & \multicolumn{1}{c}{\underline{0.186}}                       & \underline{0.331}                      & \multicolumn{1}{c}{0.188}                     & 0.333                     & \multicolumn{1}{c}{2.449}                        & 1.400                          & \multicolumn{1}{c}{0.422}                        & 0.419                        & \multicolumn{1}{c}{3.379}                       & 1.338                       & \multicolumn{1}{c}{1.193}                      & 0.884                     \\ \hline
\end{tabular}
}
\end{table*}

\subsection{Ablation Studies}
We conduct ablation studies to verify the impact of different components on long-range time series forecasting. All ablation studies are conducted on \textit{ETTh1} dataset.

\textbf{Multi-Scale Feature Extraction.} Our multi-scale feature extraction is implemented by 1D convolution. To investigate the effect of the multi-scale feature extraction, we conduct ablation studies by carefully designing the following two variants:
\begin{itemize}
\item {MSHyper-avg: It replaces 1D convolution with avg-pooling to extract multi-scale features.}
\item{MSHyper-max: It replaces 1D convolution with max-pooling to extract multi-scale features.}
\end{itemize}

The experimental results are shown in Table \ref{tab:table5}, from which we can observe that MSHyper achieves the best performance in almost all cases, which indicates that convolution kernels can extract more complex features.

\begin{table}[!t]
\centering
\caption{Results of different multi-scale feature extraction methods.}
\label{tab:table5}
\resizebox{0.34\textwidth}{!}{
\begin{tabular}{lc|cccc}
\hline
\multicolumn{2}{c|}{\multirow{2}{*}{Methods}}                                        & \multicolumn{4}{c}{Predicition Length}                                                                                                                                                                                                                                                                                 \\ \cline{3-6} 
\multicolumn{2}{c|}{}                                                                & \multicolumn{1}{c}{96}                                                    & \multicolumn{1}{c}{192}                                                            & \multicolumn{1}{c}{336}                                                            & 720                                                            \\ \hline
\multicolumn{1}{l|}{MSHyper-max} & \begin{tabular}[c]{@{}c@{}}MSE\\ MAE\end{tabular} & \multicolumn{1}{c}{\begin{tabular}[c]{@{}c@{}}0.384\\ 0.393\end{tabular}} & \multicolumn{1}{c}{\begin{tabular}[c]{@{}c@{}}0.437\\ 0.426\end{tabular}}          & \multicolumn{1}{c}{\begin{tabular}[c]{@{}c@{}}0.483\\ 0.446\end{tabular}}          & \begin{tabular}[c]{@{}c@{}}0.487\\ \textbf{0.464}\end{tabular}          \\ \hline
\multicolumn{1}{l|}{MSHyper-avg} & \begin{tabular}[c]{@{}c@{}}MSE\\ MAE\end{tabular} & \multicolumn{1}{c}{\begin{tabular}[c]{@{}c@{}}\textbf{0.382}\\ 0.393\end{tabular}} & \multicolumn{1}{c}{\begin{tabular}[c]{@{}c@{}}0.436\\ 0.426\end{tabular}}          & \multicolumn{1}{c}{{\begin{tabular}[c]{@{}c@{}}\textbf{0.478}\\ 0.447\end{tabular}}} & \begin{tabular}[c]{@{}c@{}}0.485\\ 0.468\end{tabular}          \\ \hline
\multicolumn{1}{l|}{MSHyper}     & \begin{tabular}[c]{@{}c@{}}MSE\\ MAE\end{tabular} & \multicolumn{1}{c}{\begin{tabular}[c]{@{}c@{}}0.383\\ \textbf{0.392}\end{tabular}} & \multicolumn{1}{c}{\textbf{\begin{tabular}[c]{@{}c@{}}0.435\\ 0.423\end{tabular}}} & \multicolumn{1}{c}{\begin{tabular}[c]{@{}c@{}}0.480\\ \textbf{0.445}\end{tabular}}          & {\begin{tabular}[c]{@{}c@{}}\textbf{0.482}\\ 0.467\end{tabular}} \\ \hline
\end{tabular}
}
\end{table}


\textbf{Multi-Scale Hypergraph.} To investigate the effectiveness of the multi-scale hypergraph, we conduct ablation studies by carefully designing the following three variants: 
\begin{itemize}
\item {MSHyper-FCG: It replaces the H-HGC model with the fully-connected graph.}
\item {MSHyper-PG: It replaces the H-HGC model with the pyramid graph used in Pyraformer \cite{pyraformer}.}
\item {MSHyper-SSH: It connects the input sequence with intra-scale hyperedges, and thus the multi-scale hypergraph turns to the single-scale hypergraph.}
\end{itemize}

 The experimental results are shown in Table \ref{tab:table22}, from which we can observe that MSHyper performs the best in all cases, which indicates the importance of multi-scale hypergraph in modeling high-order interactions between temporal patterns of different scales.

\begin{table}[]
\centering
\caption{Results of different multi-scale hypergraph construction methods.}
\resizebox{0.34\textwidth}{!}{
\label{tab:table22}

\begin{tabular}{lc|cccc}
\hline
\multicolumn{2}{c|}{\multirow{2}{*}{Methods}}                                        & \multicolumn{4}{c}{Predicition Length}                                                                                                                                                                                                                                                                                          \\ \cline{3-6} 
\multicolumn{2}{c|}{}                                                                & \multicolumn{1}{c}{96}                                                             & \multicolumn{1}{c}{192}                                                            & \multicolumn{1}{c}{336}                                                            & 720                                                            \\ \hline
\multicolumn{1}{l|}{MSHyper-FCG} & \begin{tabular}[c]{@{}c@{}}MSE\\ MAE\end{tabular} & \multicolumn{1}{c}{\begin{tabular}[c]{@{}c@{}}0.699\\ 0.562\end{tabular}}          & \multicolumn{1}{c}{\begin{tabular}[c]{@{}c@{}}0.723\\ 0.574\end{tabular}}          & \multicolumn{1}{c}{\begin{tabular}[c]{@{}c@{}}0.732\\ 0.583\end{tabular}}          & \begin{tabular}[c]{@{}c@{}}0.740\\ 0.622\end{tabular}          \\ \hline
\multicolumn{1}{l|}{MSHyper-PG}  & \begin{tabular}[c]{@{}c@{}}MSE\\ MAE\end{tabular} & \multicolumn{1}{c}{\begin{tabular}[c]{@{}c@{}}0.453\\ 0.449\end{tabular}}          & \multicolumn{1}{c}{\begin{tabular}[c]{@{}c@{}}0.494\\ 0.471\end{tabular}}          & \multicolumn{1}{c}{\begin{tabular}[c]{@{}c@{}}0.532\\ 0.493\end{tabular}}          & \begin{tabular}[c]{@{}c@{}}0.534\\ 0.551\end{tabular}          \\ \hline
\multicolumn{1}{l|}{MSHyper-SSH} & \begin{tabular}[c]{@{}c@{}}MSE\\ MAE\end{tabular} & \multicolumn{1}{c}{\begin{tabular}[c]{@{}c@{}}0.432\\ 0.427\end{tabular}}          & \multicolumn{1}{c}{\begin{tabular}[c]{@{}c@{}}0.453\\ 0.461\end{tabular}}          & \multicolumn{1}{c}{\begin{tabular}[c]{@{}c@{}}0.510\\ 0.470\end{tabular}}          & \begin{tabular}[c]{@{}c@{}}0.518\\ 0.491\end{tabular}          \\ \hline
\multicolumn{1}{l|}{MSHyper}     & \begin{tabular}[c]{@{}c@{}}MSE\\ MAE\end{tabular} & \multicolumn{1}{c}{\textbf{\begin{tabular}[c]{@{}c@{}}0.383\\ 0.392\end{tabular}}} & \multicolumn{1}{c}{\textbf{\begin{tabular}[c]{@{}c@{}}0.435\\ 0.423\end{tabular}}} & \multicolumn{1}{c}{\textbf{\begin{tabular}[c]{@{}c@{}}0.480\\ 0.445\end{tabular}}} & \textbf{\begin{tabular}[c]{@{}c@{}}0.482\\ 0.467\end{tabular}} \\ \hline
\end{tabular}
}
\end{table}

\textbf{Hyperedge Graph.} To investigate the effectiveness of the hyperedge graph, we conduct ablation studies by carefully designing the following three variants: 
\begin{itemize}
\item { MSHyper-w/o AR: It removes the association relationship of the hyperedge graph.}
\item {MSHyper-w/o SR: It removes the sequential relationship of the hyperedge graph.}
\item {MSHyper-w/o H-H: It removes the hyperedge graph (i.e., without the hyperedge-hyperedge phase).}
\end{itemize}

The experimental results are shown in Table \ref{tab:table4}, from which we can discern the following tendencies:
1) MSHyper performs better than MSHyper-w/o AR and MSHyper-w/o SR, showing the effectiveness of the association relationship and sequential relationship, respectively. 2) Removing the hyperedge-hyperedge phase gets the worst forecasting results, which demonstrates the superiority of the hyperedge graph in enhancing hypergraph modeling.

\begin{table}[!t]
\centering
\caption{Results of different hyperedge graph construction methods.}
\label{tab:table4}
\resizebox{0.34\textwidth}{!}{
\begin{tabular}{cc|cccc}
\hline
\multicolumn{2}{c|}{\multirow{2}{*}{Methods}}                                            & \multicolumn{4}{c}{Predicition Length}                                                                                                                                                                                                                                                                                          \\ \cline{3-6} 
\multicolumn{2}{c|}{}                                                                    & \multicolumn{1}{c}{96}                                                             & \multicolumn{1}{c}{192}                                                            & \multicolumn{1}{c}{336}                                                            & 720                                                            \\ \hline
\multicolumn{1}{l|}{MSHyper-w/o AR}  & \begin{tabular}[c]{@{}c@{}}MSE\\ MAE\end{tabular} & \multicolumn{1}{c}{\begin{tabular}[c]{@{}c@{}}0.391\\ 0.399\end{tabular}}          & \multicolumn{1}{c}{\begin{tabular}[c]{@{}c@{}}0.441\\ 0.429\end{tabular}}          & \multicolumn{1}{c}{\begin{tabular}[c]{@{}c@{}}0.499\\ 0.486\end{tabular}}          & \begin{tabular}[c]{@{}c@{}}0.503\\ 0.493\end{tabular}          \\ \hline
\multicolumn{1}{l|}{MSHyper-w/o SR}  & \begin{tabular}[c]{@{}c@{}}MSE\\ MAE\end{tabular} & \multicolumn{1}{c}{\begin{tabular}[c]{@{}c@{}}0.385\\ 0.393\end{tabular}}          & \multicolumn{1}{c}{\textbf{\begin{tabular}[c]{@{}c@{}}0.435\\ 0.422\end{tabular}}} & \multicolumn{1}{c}{\begin{tabular}[c]{@{}c@{}}0.487\\ 0.453\end{tabular}}          & \begin{tabular}[c]{@{}c@{}}0.496\\ 0.487\end{tabular}          \\ \hline
\multicolumn{1}{l|}{MSHyper-w/o H-H} & \begin{tabular}[c]{@{}c@{}}MSE\\ MAE\end{tabular} & \multicolumn{1}{c}{\begin{tabular}[c]{@{}c@{}}0.433\\ 0.434\end{tabular}}          & \multicolumn{1}{c}{\begin{tabular}[c]{@{}c@{}}0.479\\ 0.459\end{tabular}}          & \multicolumn{1}{c}{\begin{tabular}[c]{@{}c@{}}0.519\\ 0.477\end{tabular}}          & \begin{tabular}[c]{@{}c@{}}0.524\\ 0.498\end{tabular}          \\ \hline
\multicolumn{1}{l|}{MSHyper}         & \begin{tabular}[c]{@{}c@{}}MSE\\ MAE\end{tabular} & \multicolumn{1}{c}{\textbf{\begin{tabular}[c]{@{}c@{}}0.383\\ 0.392\end{tabular}}} & \multicolumn{1}{c}{{\begin{tabular}[c]{@{}c@{}}\textbf{0.435}\\ 0.423\end{tabular}}} & \multicolumn{1}{c}{\textbf{\begin{tabular}[c]{@{}c@{}}0.480\\ 0.445\end{tabular}}} & \textbf{\begin{tabular}[c]{@{}c@{}}0.482\\ 0.467\end{tabular}} \\ \hline
\end{tabular}
}
\end{table}
\subsection{Case Studies}
\begin{figure*}[]
    \centering
    \begin{minipage}[t]{0.24\linewidth}
    	\centering	
        \includegraphics[width=1\linewidth]{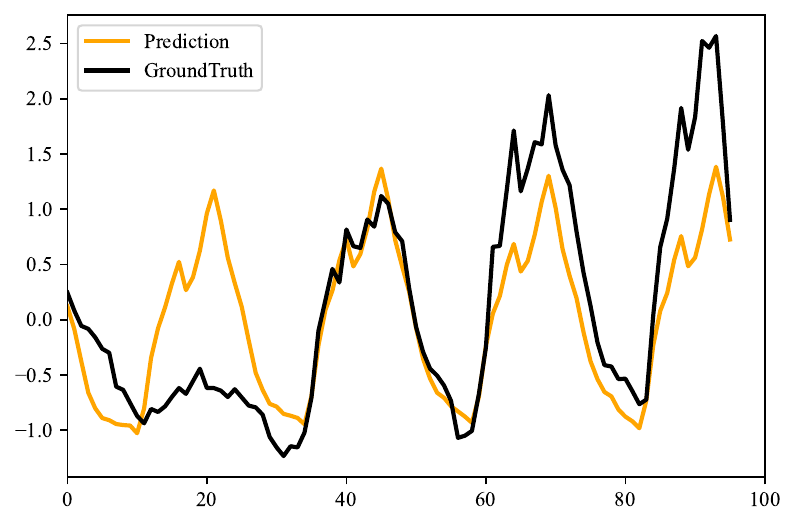}
        \put(-60,-10){{\makebox(0,0)[b]{\fontsize{5}{6} \selectfont (a) DLinear}}}
    \end{minipage}%
    \hspace{0.01\linewidth}
    \begin{minipage}[t]{0.24\linewidth}
        \centering
        \includegraphics[width=1\linewidth]{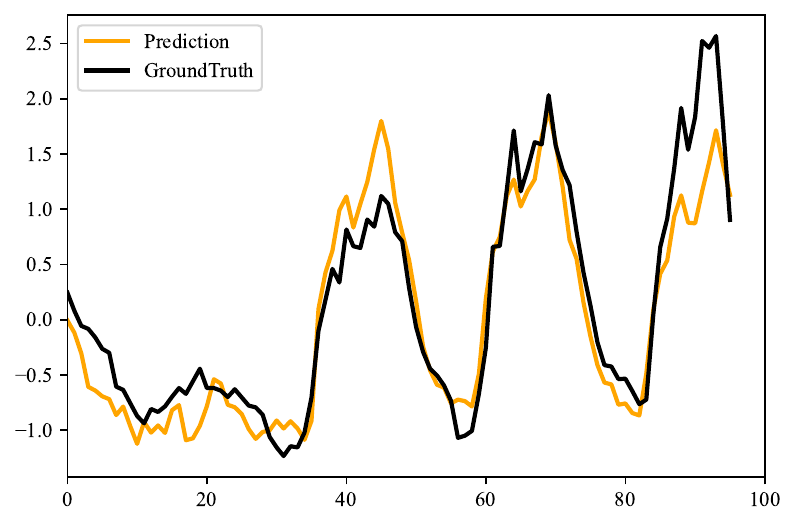}
        \put(-60,-10){{\makebox(0,0)[b]{\fontsize{5}{6} \selectfont (b) TimesNet}}}
    \end{minipage}%
    \begin{minipage}[t]{0.24\linewidth}
        \centering
        \includegraphics[width=1\linewidth]{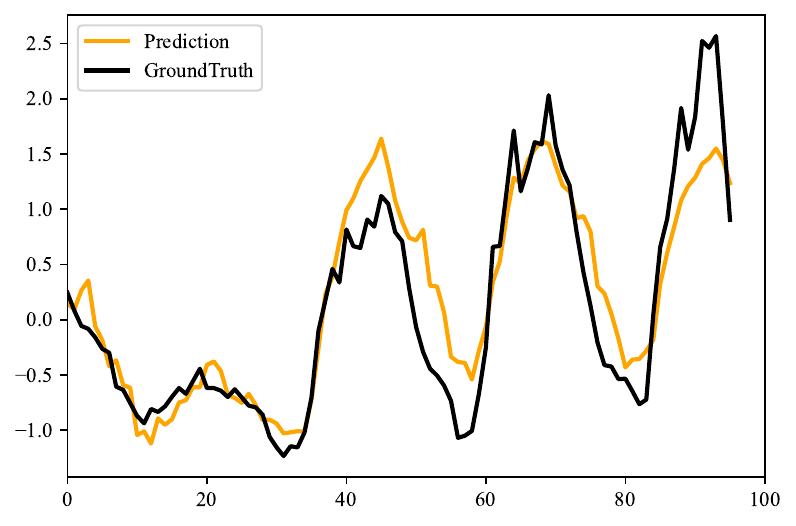}
        \put(-60,-10){{\makebox(0,0)[b]{\fontsize{5}{6} \selectfont (c) MSGNet}}}
    \end{minipage}%
    \begin{minipage}[t]{0.24\linewidth}
        \centering
        \includegraphics[width=1\linewidth]{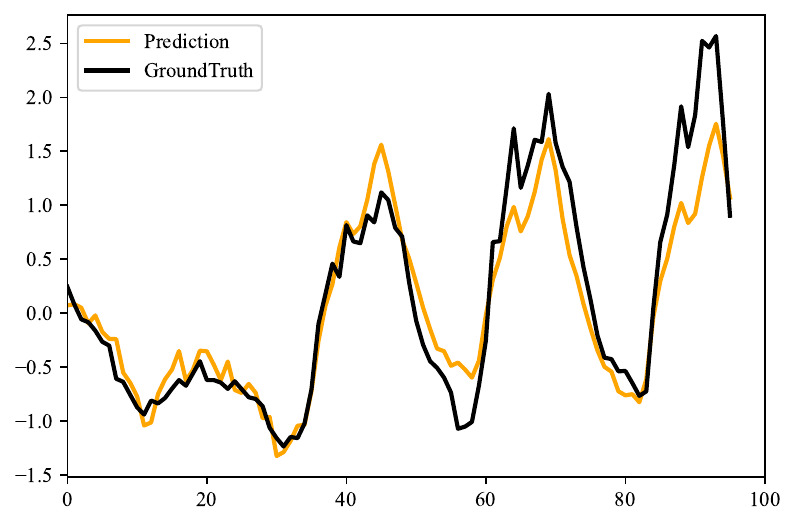}
        \put(-60,-10){{\makebox(0,0)[b]{\fontsize{5}{6} \selectfont (d) MSHyper}}}
    \end{minipage}%
    \caption{Forecasting results of different models on \textit{Electricity} dataset under the input-96-predict-96 setting. The black line represents the ground truth and the orange line represents the predicted results.}
    \label{Figure_5}  
\end{figure*}

\begin{figure*}[]
    \centering
    \begin{minipage}[t]{0.24\linewidth}
    	\centering	
        \includegraphics[width=1\linewidth]{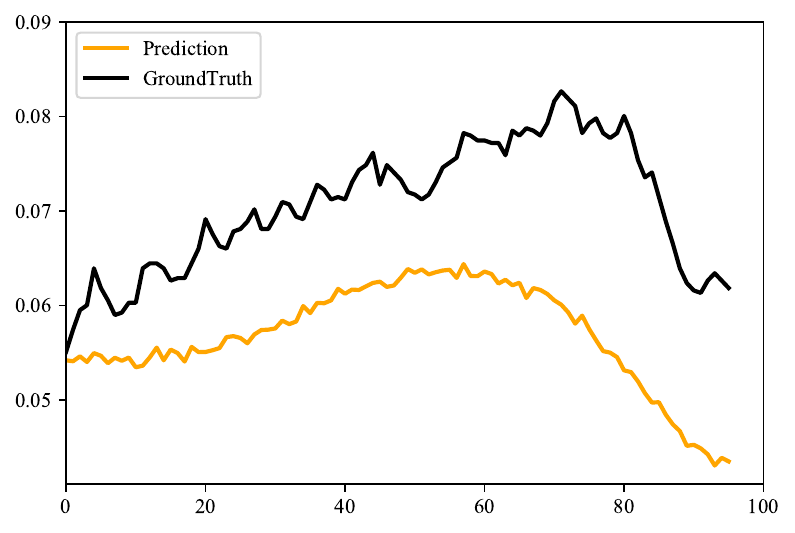}
        \put(-60,-10){{\makebox(0,0)[b]{\fontsize{5}{6} \selectfont (a) DLinear}}}
    \end{minipage}%
    \hspace{0.01\linewidth}
    \begin{minipage}[t]{0.24\linewidth}
        \centering
        \includegraphics[width=1\linewidth]{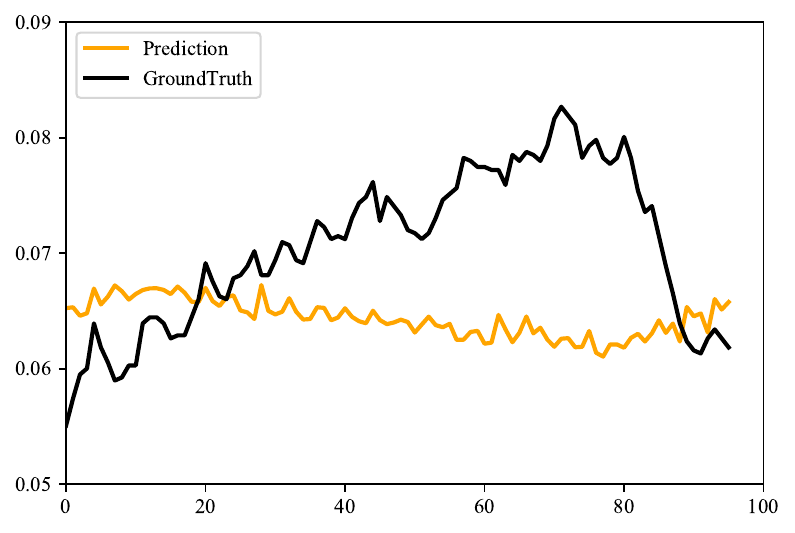}
        \put(-60,-10){{\makebox(0,0)[b]{\fontsize{5}{6} \selectfont (b) TimesNet}}}
    \end{minipage}%
    \begin{minipage}[t]{0.24\linewidth}
        \centering
        \includegraphics[width=1\linewidth]{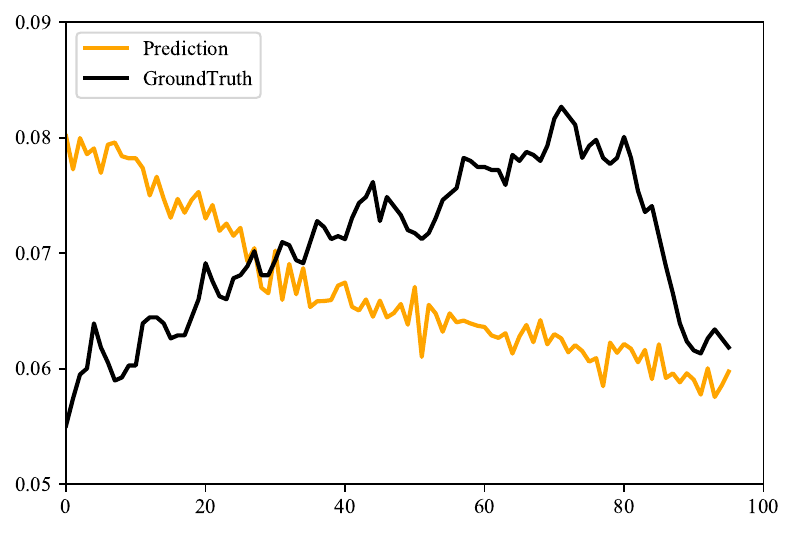}
        \put(-60,-10){{\makebox(0,0)[b]{\fontsize{5}{6} \selectfont (c) MSGNet}}}
    \end{minipage}%
    \begin{minipage}[t]{0.24\linewidth}
        \centering
        \includegraphics[width=1\linewidth]{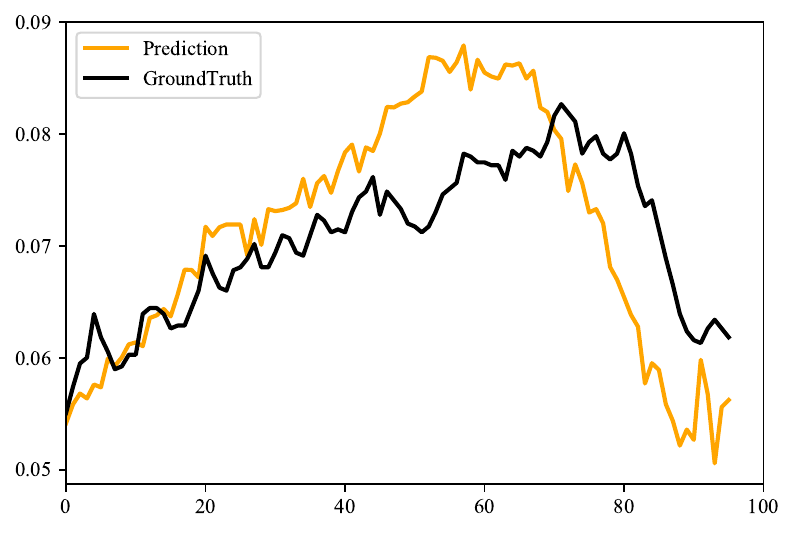}
        \put(-60,-10){{\makebox(0,0)[b]{\fontsize{5}{6} \selectfont (d) MSHyper}}}
    \end{minipage}%
    \caption{Forecasting results of different models on \textit{Weather} dataset under the input-96-predict-96 setting. The black line represents the ground truth and the orange line represents the predicted results.}
    \label{Figure_6}  
\end{figure*}

\begin{figure}[!t]
    \centering
    \begin{minipage}[t]{0.45\linewidth}
    	\centering	
        \includegraphics[width=1\linewidth]{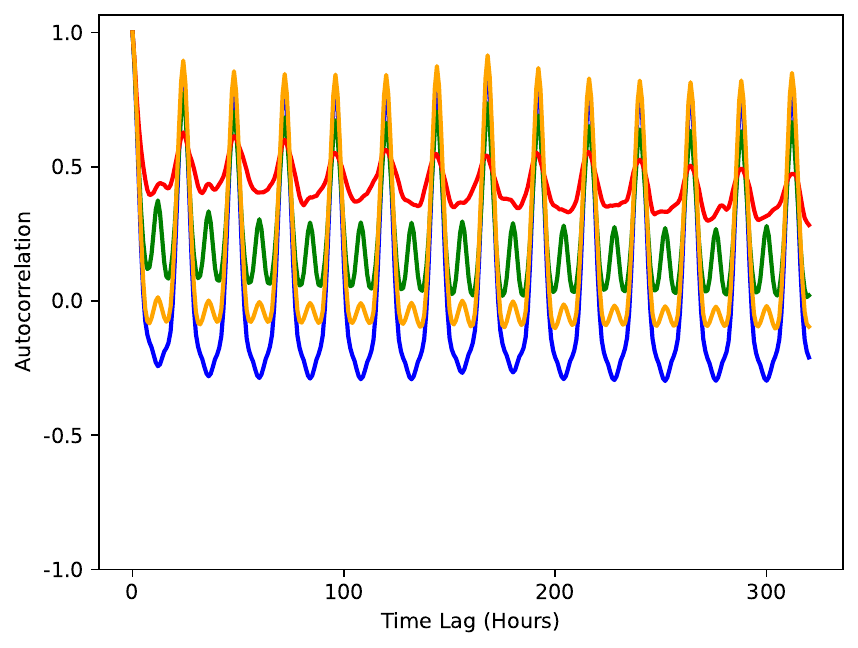}
        \put(-55,-10){{\makebox(0,0)[b]{\fontsize{5}{6} \selectfont (a) Electricity dataset}}}
    \end{minipage}%
    \hspace{0.01\linewidth}
    \begin{minipage}[t]{0.45\linewidth}
        \centering
        \includegraphics[width=1\linewidth]{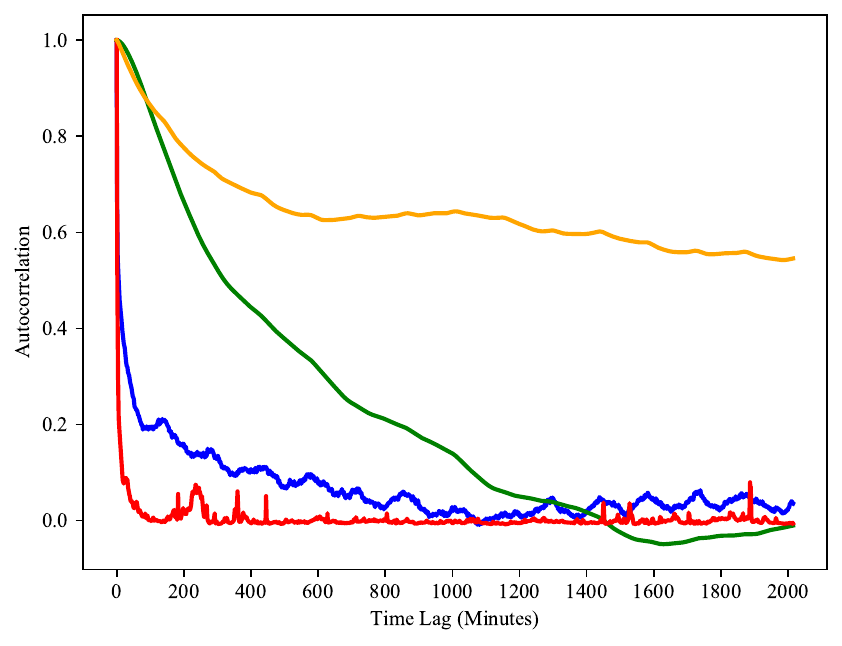}
        \put(-55,-10){{\makebox(0,0)[b]{\fontsize{5}{6} \selectfont (b) Weather dataset}}}
    \end{minipage}%
    \caption{The auto-correlation graphs on \textit{Electricity} and \textit{Weather} datasets.}
    \label{Figure_7}  
\end{figure}

To evaluate the prediction performance of different models, we plot the forecasting results on \textit{Weather} and \textit{Electricity} datasets for qualitative comparison. The visualization of forecasting results are shown in Figure \ref{Figure_5} and \ref{Figure_6}, from which we can observe that the existing models have poorer fitting ability on \textit{Weather} dataset than on \textit{Electricity} dataset. To explore the reason, Figure \ref{Figure_7} shows the auto-correlation graphs of sampled variables on \textit{Electricity} and \textit{Weather} datasets. On \textit{Electricity} dataset, clear daily and weekly patterns can be observed, while on \textit{Weather} dataset, it is difficult to identify obvious daily or weekly patterns. In addition, MSHyper is still able to achieve good predictive performance on \textit{Weather} dataset, which may be due to its ability to capture more diverse long-term and short-term patterns by modeling high-order temporal pattern interactions.

\subsection{Parameter Studies}

\begin{figure}[]
    \centering
    \begin{minipage}[t]{0.40\linewidth}
    	\centering	
        \includegraphics[width=1\linewidth]{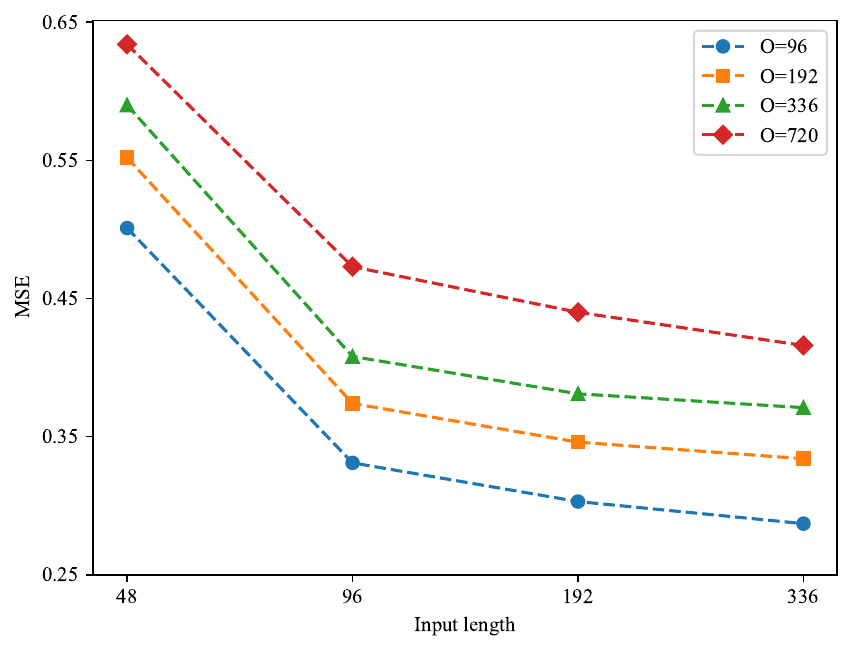}
        \put(-51,-5){{\makebox(0,0)[b]{\fontsize{4}{5} \selectfont Input length}}}
        \put(-102,36){\rotatebox{90}{\makebox(0,0)[b]{\fontsize{4}{5} \selectfont{MSE}}}}
    \end{minipage}%
    \hspace{0.01\linewidth}
    \begin{minipage}[t]{0.40\linewidth}
        \centering
        \includegraphics[width=1\linewidth]{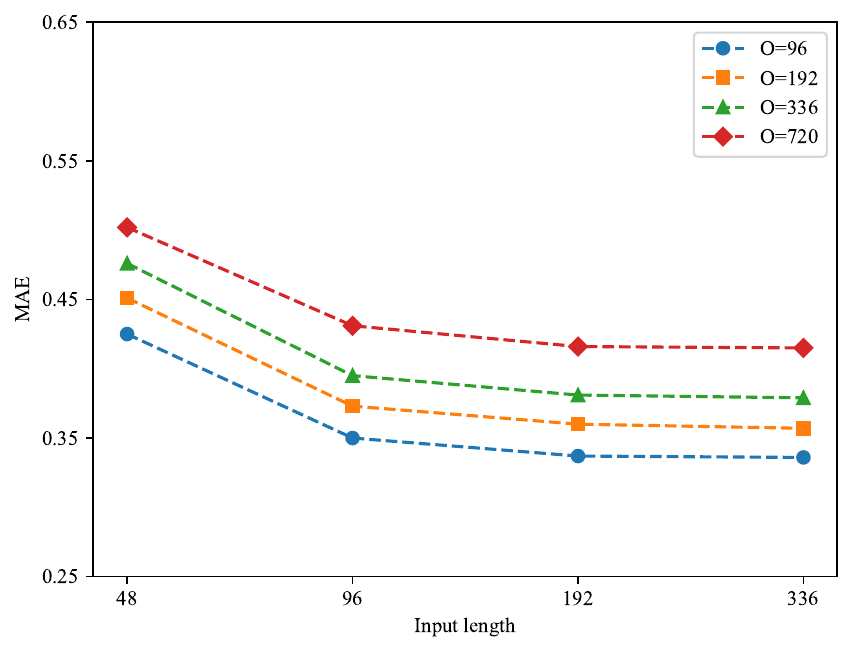}
        \put(-50,-5){{\makebox(0,0)[b]{\fontsize{4}{5} \selectfont Input length}}}
        \put(-102,36){\rotatebox{90}{\makebox(0,0)[b]{\fontsize{4}{5} \selectfont{MAE}}}}
    \end{minipage}%
    \caption{Results of MSHyper with different input length on \textit{ETTm1} dataset.}
    \label{fig4}  
\end{figure}

\textbf{Input Length.} To investigate the impact of input length,  we set input length $T\in\{48, 96, 192, 336\}$ and record the results of MSHyper for multivariate long-range time series forecasting. The experimental results are shown in Figure \ref{fig4}, from which we can observe that with the increase of input length, the performance of MSHyper shows an upward trend. The reason is that MSHyper can capture more historical pattern information with the increase of input length.

\begin{figure}[!t]
    \centering
    \begin{minipage}[t]{0.40\linewidth}
    	\centering	
        \includegraphics[width=1\linewidth]{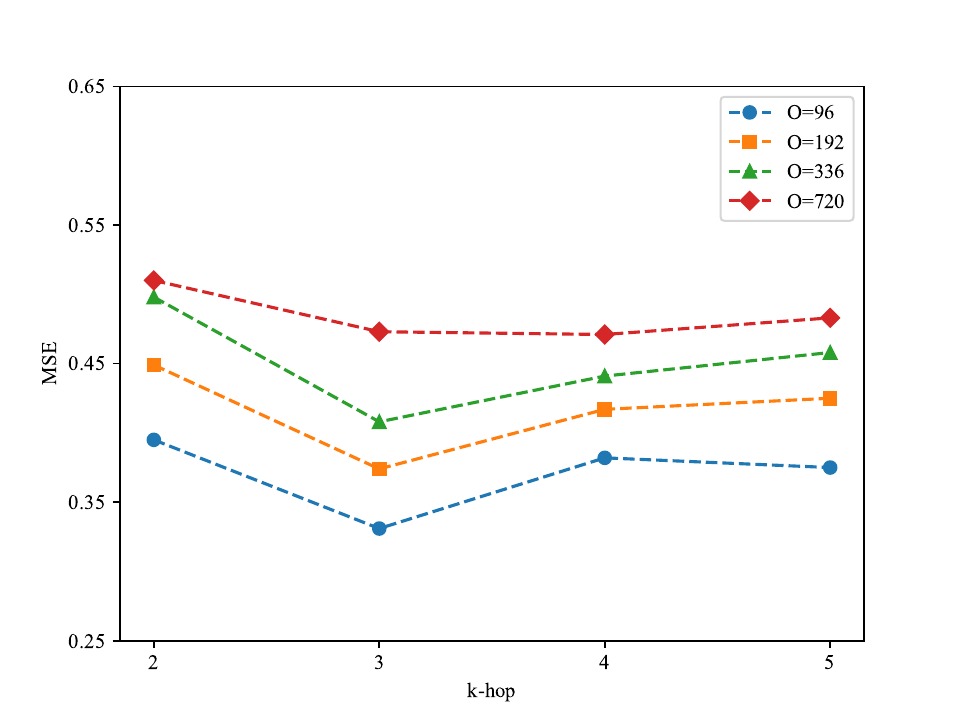}
        \put(-50,-5){{\makebox(0,0)[b]{\fontsize{4}{5} \selectfont \textit{k}-hop}}}
        \put(-102,36){\rotatebox{90}{\makebox(0,0)[b]{\fontsize{4}{5} \selectfont{MSE}}}}
    \end{minipage}%
    \hspace{0.01\linewidth}
    \begin{minipage}[t]{0.40\linewidth}
        \centering
        \includegraphics[width=1\linewidth]{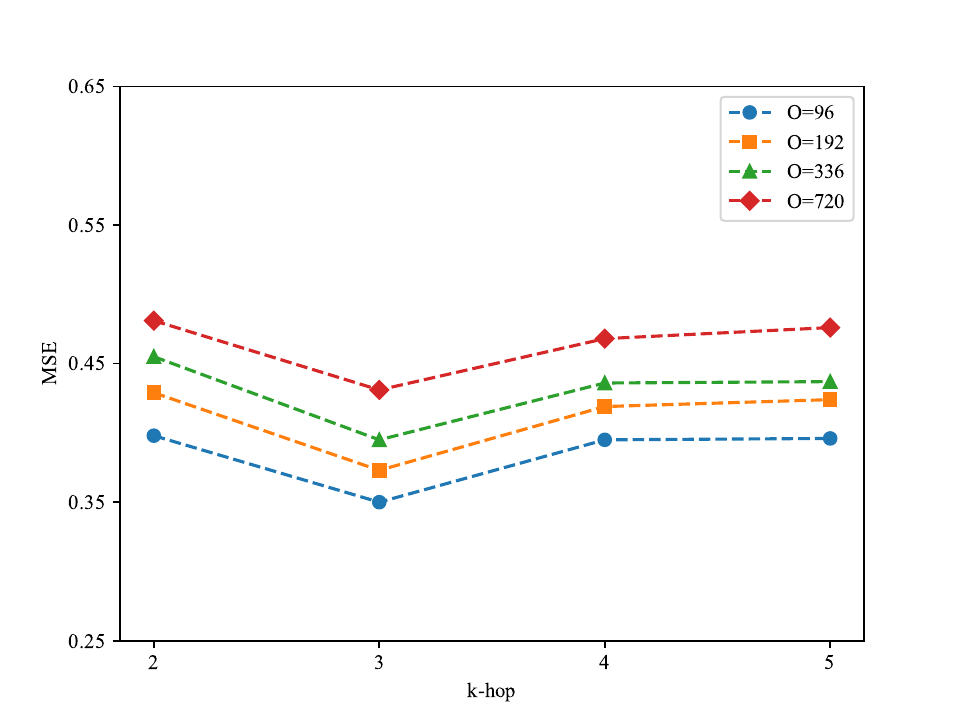}
        \put(-50,-5){{\makebox(0,0)[b]{\fontsize{4}{5} \selectfont \textit{k}-hop}}}
        \put(-102,36){\rotatebox{90}{\makebox(0,0)[b]{\fontsize{4}{5} \selectfont{MAE}}}}
    \end{minipage}%
    \caption{Results of MSHyper with different $k$-hop on \textit{ETTm1} dataset.}
    \label{fig5}  
\end{figure}

\textbf{$\boldsymbol{k}$-hop.} We also perform parameter studies to measure the impact of the $k$-hop on \textit{ETTm1} dataset. Figure \ref{fig5} shows the results of MSHyper on \textit{ETTm1} dataset by varying the $k$-hop from 2 to 5, form which we can observe that the best performance can be obtained when the $k$-hop is 3. The reason is that a small $k$-hop limits interactions to the neighboring nodes, and a large $k$-hop would introduce noises.

\subsection{Computation Cost}
To evaluate the computation cost of MSHyper, we compare the training time, GPU occupation, and forecasting performance of MSHyper, MSGNet, TimesNet, Crossformer, and DLinear on \textit{Electricity} dataset with the forecasting horizon of 96. The experimental results are shown in Table \ref{tab:table7}, from which we can observe that DLinear has lower GPU occupation and runs fastest in these methods. But it gets worse forecasting performance than MSHyper, MSGNet, and TimesNet. Compared with MSGNet, TimesNet, and Crossformer, MSHyper runs faster and gets the best forecasting performance. Overall, comprehensively considering the significant forecasting performance improvement and the computation cost, MSHyper demonstrates the superiority over existing methods.
\begin{table}[!t]
\centering
\caption{Computation costs of different methods.}
\label{tab:table7}
\resizebox{0.45\textwidth}{!}{
\begin{tabular}{lllll}
\hline
Methods     & Training Time/epoch  & GPU Occupation & MSE            & MAE            \\ \hline
MSHyper     & 13.51s         & 2186           & \textbf{0.152} & \textbf{0.252} \\ 
MSGNet      & 1519.10s       & 11550          & 0.165          & 0.274          \\ 
TimesNet    & 587.35s        & 3726           & 0.168          & 0.272          \\ 
Crossformer & 88.36s         & 9468           & 0.219          & 0.314          \\ 
DLinear     & \textbf{4.35s} & \textbf{756}   & 0.197          & 0.282          \\ \hline
\end{tabular}
}
\end{table}

\section{Conclusions and Future Work}
In this paper, we propose MSHyper for long-range time series forecasting. Specifically, the H-HGC module is introduced to provide foundations for modeling high-order interactions between temporal patterns. The TMP mechanism is employed to aggregate high-order pattern information and learn the interaction strength between temporal patterns of different scales. Extensive experiments on eight real-world datasets show the superiority of MSHyper.

In future work, it is of interest to extend MSHyper in the following two aspects: First, since the prior knowledge-based hypergraph structures may limit the scalability of MSHyper, we will design an adaptive hypergraph module to learn the best hypergraph structures. Second, to tackle complex scenariors with diverse temporal pattern interactions, we will design a neural architecture search framework to search the best combination of hypergraph structures, capturing high-order correlations among temporal patterns of different scales automatically.
\bibliographystyle{plain}
\bibliography{TKDE24}

\vfill

\end{document}